\documentclass{article}

\usepackage[nonatbib, preprint]{neurips_2024}
\usepackage[square,sort,comma,numbers]{natbib}

\usepackage[utf8]{inputenc} 
\usepackage[T1]{fontenc}    
\usepackage{hyperref}       
\usepackage{url}            
\usepackage{booktabs}       
\usepackage{amsfonts}       
\usepackage{nicefrac}       
\usepackage{microtype}      
\usepackage{xcolor}         

\usepackage{bm}
\usepackage{amsmath}
\usepackage{xspace}
\usepackage{graphicx}
\usepackage{titletoc}
\usepackage{enumitem}
\usepackage{bbm}
\usepackage{lmodern}
\usepackage{wrapfig}
\newcommand{\method}{\texttt{TrialSynth}\xspace}
\newcommand\DoToC{%
  \startcontents
  \printcontents{}{2}{\textbf{Contents}\vskip3pt\hrule\vskip5pt}
  \vskip3pt\hrule\vskip5pt
}

\title{\method: Generation of Synthetic Sequential Clinical Trial Data}

\author{%
  Chufan Gao \\
  University of Illinois Urbana Champaign\\
  \texttt{chufan2@illinois.edu} \\
  \And
  Mandis Beigi \\
  Medidata Solutions \\
  \texttt{mandis.beigi@3ds.com} \\
  \And
  Afrah Shafquat \\
  Medidata Solutions \\
  \texttt{afrah.shafquat@3ds.com} \\
  \And
  Jacob Aptekar \\
  Medidata Solutions \\
  \texttt{jacob.aptekar@3ds.com} \\
  \And
  Jimeng Sun \\
  University of Illinois Urbana Champaign\\
  Carle Illinois College of Medicine \\
  \texttt{jimeng@illinois.edu} \\
}

\begin{document}

\maketitle

\begin{abstract}
Analyzing data from past clinical trials is part of the ongoing effort to optimize the design, implementation, and execution of new clinical trials and more efficiently bring life-saving interventions to market.
While there have been recent advances in the generation of static context synthetic clinical trial data, due to both limited patient availability and constraints imposed by patient privacy needs, the generation of fine-grained synthetic time-sequential clinical trial data has been challenging. Given that patient trajectories over an entire clinical trial are of high importance for optimizing trial design and efforts to prevent harmful adverse events, there is a significant need for the generation of high-fidelity time-sequence clinical trial data. Here we introduce \method, a Variational Autoencoder (VAE) designed to address the specific challenges of generating synthetic time-sequence clinical trial data.  Distinct from related clinical data VAE methods, the core of our method leverages Hawkes Processes (HP), which are particularly well-suited for modeling event-type and time gap prediction needed to capture the structure of sequential clinical trial data. Our experiments demonstrate that \method surpasses the performance of other comparable methods that can generate sequential clinical trial data at varying levels of fidelity / privacy tradeoff, enabling the generation of highly accurate event sequences across multiple real-world sequential event datasets with small patient source populations. Notably, our empirical findings highlight that \method not only outperforms existing clinical sequence-generating methods but also produces data with superior utility while empirically preserving patient privacy. 
\end{abstract}

\section{Introduction}

The data generated from past clinical trials represent a valuable resource for informing drug development \cite{fu2022hint,chen2024uncertainty, wang2023pytrial,lu2024uncertainty} and increasing the speed at which vital life-saving drugs arrive to market \cite{friedman2015fundamentals, downing2014clinical} While the potential value of clinical trial data is high, these data are often inaccessible due to patient privacy concerns and legal constraints \citep{ shafquat2023interpretable,lu2018multi,lu2024drugclip}. The generation of high-quality synthetic clinical trial data that captures the properties of real data while simultaneously protecting patient privacy is increasingly being seen as a strategy for sharing and applying these data in drug development applications \cite{james2021synthetic}.  

Though proposed methods for generating synthetic clinical trial data have focused on static context information for each subject (e.g., demographics) \cite{hernandez2022synthetic}, many of the highest value applications, including control arm augmentation \cite{thorlund2020synthetic} require generating synthetic time-sequential event data that has high fidelity \citep{beigi2022simulants,yi2018enhance}. However, developing a high-quality model for sequential trial data can be more complicated than data-rich tasks in computer vision or natural language processing, due to the small sample size of training datasets available, which is less common in other applications of generative models.

To address these challenges, we propose \method. This method makes use of Hawkes processes, which are statistical models that are specialized for event-type and time gap prediction \citep{hawkes1971spectra, zuo2020transformer}, as well as Variational Autoencoders (VAEs) \citep{kingma2013auto}, a proven generative framework that has worked well for static clinical trial data synthesis \cite{das2023twin}. We empirically demonstrate that combining these two classical approaches leads to an algorithm that is capable of generating sequential event synthetic data even on small amounts of clinical trial data.  

\begin{figure*}
    \centering
    \includegraphics[width=.9\linewidth]{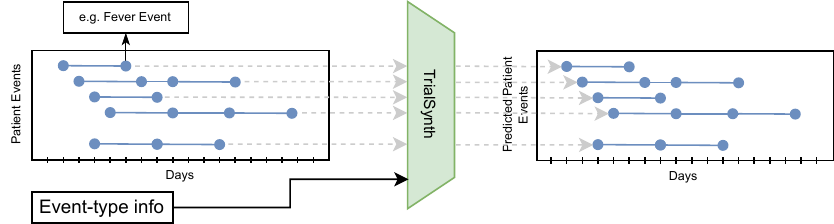}
    \caption{Visualization of data input and synthetic data generation of \method, the model input is the real patient events and their timestamps, and we wish to generate synthetic patient events and their timestamps. This is a particularly challenging task due to the small amount of patient data. \method also explicitly supports adding the event type information in the form of specifying the specific event types to generate.}
    \label{fig:data_intro}
    \vspace{-1em}
\end{figure*}
To summarize our contributions:
\begin{enumerate}[leftmargin=*]
    \item We introduce \method--a model that combines Variational Autoencoder + Hawkes Process that is both able to generate sequential event clinical trial data and supports a high level of control, allowing users to specify specific event types and variance levels to generate (\url{https://github.com/chufangao/TrialSynth}). 
    \item We demonstrate from the analysis of 7 real-world clinical trial datasets that \method outperforms alternative approaches designed for tabular data generation. 
    \item We also demonstrate \method achieves high performance versus privacy trade-off with two key metrics: ML Inference Score, which shows that synthetic event sequences are hard to distinguish from the original sequences, and Distance to Closest Record (DCR), which shows that synthetic sequences are not copies of the original data.
\end{enumerate}

The rest of this paper is organized as follows: In Section~\ref{sec:Related Work}, we review the related work. In Section~\ref{sec:method}, we dive into the proposed \method
in detail. In section~\ref{sec:Experiments}, we compare datasets and baselines, demonstrating the superiority of \method.
Finally, in Section~\ref{sec:Discussion}, we provide a discussion and conclude our findings.

\section{Related Work}  
\label{sec:Related Work}
\textbf{Synthetic Data Generation} as a research area has been quickly garnering attention from the research community, with examples such as CTGAN \citep{ctgan}, CTabGan \citep{zhao2022ctab}, TabDDPM \citep{kotelnikov2023tabddpm}, TWIN-GPT~\citep{wang2024twin}, the Synthetic Data Vault\footnote{\url{https://docs.sdv.dev/sdv/}} \citep{SDV}, and more. However, most of these models, such as TabDDPM and CTGAN, are focused on explicitly generating tabular data with no time component; or, in the case of SDV's ParSynthesizer \citep{par}, it is relatively simple and may be approximated with a GRU or LSTM model. 

\textbf{Trial Patient Generation} is a research area that has become popular.
In Electronic Healthcare Record (EHR) generation \citep{choi2017generating,chen2021data,lu2022cot, das2023twin, wang2023pytrial, lu2023machine,theodorou2023synthesize,wang2024twin}, the model usually only focuses on generating the \textit{order} at which certain clinical events happen (i.e., the diagnosis code of next patient visit), as opposed to generating the specific times of the visits as well. For example, \cite{das2023twin,wang2024twin} generates a digital twin of an input patient event sequence via a VAE and a cross-modality model, but cannot handle event timestamp generation. \method extends this line of previous work to include the specific timestamps on which these events occur, as well as the order. \cite{theodorou2023synthesize} created a strong patient EHR generation baseline, but relies on a high amount of training data (929,268 and 46,520 patients in outpatient and inpatient datasets respectively). However, in a single clinical trial, all of our datasets contain less than 1000 patients, which makes HALO difficult to run. \method is designed for and performs well on small clinical trial datasets, particularly if the event types are known. 

\textbf{Hawkes Processes combined with VAEs} is an area of research that is particularly appealing for our scenario.
We employ the Transformer Hawkes Process \cite{zuo2020transformer} for our data generation modeling. To the best of our knowledge, \method is the first to extend Hawkes models to full patient event generation from a single embedding. Unlike the Hawkes Process, it relaxes the assumption that past events can never lower the probability of future events, and performs much better on real world data.


This inherent capability of modeling events and their time occurrences makes Hawkes Processes highly suitable for event prediction. Previous work explores variational Hawkes processes in the context of event prediction for (disease progression \citep{chen2024trialbench} and social events sequences \cite{pan2020variational}, but they rely on the context of previous ground truth observations as well as the hidden state. Another work \citep{lin2021disentangled} explores using variational approaches to disentangle multivariate Hawkes Process for event type prediction, but it also relies on knowing the ground truth to predict the next timestep. This limitation is a major roadblock in a full synthetic data generation setting. Because of this, there is leaking of information from ground truth event occurrences. This information leakage is not permitted in our task, which is a fully generative setting from the embedding space. 


\section{\method}
\label{sec:method}
\begin{figure*}[ht]
    \centering
    \includegraphics[width=.8\linewidth]{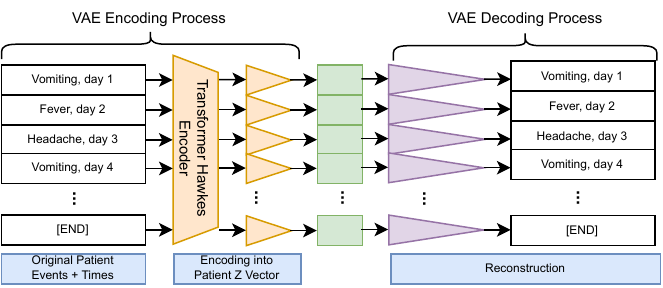}
    \caption{Diagram of the \method Encoder-Decoder structure. Here, the model input is the real patient event sequence + time, which trains a VAE model to the same output time + event sequence. The event sequence length for each event is also predicted. The transformer encoder processes each input timestep, then output embeddings are individually transformed to the z-latent space via a neural network. Sampling and decoding occur from each timestep-specific z-latent representation.}
    \label{fig:multiple}
    \vspace{-1em}
\end{figure*}
\method is created to solve the highly specific task of synthetic sequential clinical trial patient generation. As shown in Fig.~\ref{fig:data_intro} and Fig.~\ref{fig:example}, a patient contains many sequences of event types and their timestamps. This essentially creates a high-vocabulary, sequential token (event types) generation problem with a regression component (event times). First, we formulate the components that compose \method. Then, we explain key details of \method, including the ability to input type information in the form of known types to generate (which is common in the trial generation space when up-sampling patient data). Finally, we conclude with experiments on ML utility (usefulness of synthetic data) and inference privacy (important for patient privacy) and a discussion of the results.

\subsection{Encoding and Decoding Hawkes Processes}
Neural Hawkes Process \citep{mei2017neural} was proposed to generalize the traditional Hawkes Process. Let us describe the $\lambda(t)$, the intensity function of any event occurring at time $t$.
$$\lambda(t) := \sum_{k=1}^K \lambda_k(t) := \sum_{k=1}^K f_k(\bm{W}_k^\top \bm{h}(t)) = \sum_{k=1}^K \beta_k \log \left(1+ e^{\frac{\bm{W}_k^T \bm{h}(t)}{\beta_k}} \right), $$
$\lambda_k(t)$ is the intensity function for the event $k \in \mathcal{K}$ occurring, $K = |\mathcal{K}|$ is the total number of event types, $h(t)$ are the hidden states of the event sequence obtained by a Transformer encoder, and $\bm{W}_k^\top$ are learned weights that calculate the significance of each event type at time $t$. 
$f_k(c) = \beta_k \log(1+e^{\frac{x}{\beta_k}})$ is the softplus function with parameter $\beta_k$. The output of $f_k(x)$ is always positive. Note that the positive intensity does not mean that the influence is always positive, as the influence of previous events is calculated through $\bm{W}_k^\top \bm{h}(t)$. If there is an event occurring at time $t$, then the probability of event $k$ is $P(k_t=k) = \frac{\lambda_k(t)}{\lambda(t)}$.
Furthermore, the log-likelihood is:
$\ln P_\theta(\{(t_1, k_1), \dots, (t_{L}, k_{L})\}|\bm{z}) = \sum_{j=1}^{L} \log(\lambda_\theta(t_{j}| \mathcal{H}_{t_{j},z})) - \int_{t_1}^{t_{L}} \lambda_\theta(t | \mathcal{H}_{t,z})  dt. $

\textbf{Encoder:} The encoder model $E_{\method}(\mathcal{H}_i) \rightarrow \hat{\bm{\mu}}, \hat{\bm{\sigma}}$ takes in the original event types and times, and predicts the mean and standard deviation to sample hidden state vector at time $\bm{z_t}$ at each timestep $t$. These $\bm{z_t}$ are concatenated to form $\bm{z} \sim Normal(\hat{\bm{\mu}}, \hat{\bm{\sigma}})$. $\bm{z}$ is trained to be close to the $Normal(\bm{0}, \bm{1})$ via ELBO. 

\textbf{Decoder:} We train the decoder to maximize the likelihood of the input Hawkes Process. I.e. the input is the ground truth event type and time-step sequence, and the autoencoder reconstructs it from $\bm{z}$. For our purposes, we adapt a decoding scheme similar to HALO \cite{theodorou2023synthesize}.

At \textit{training time}, the input to the decoder $$D_\method(\bm{z}, (t_1, k_1), \dots (t_i, k_i)) \rightarrow (\hat{t}_{i+1}, \hat{k}_{i+1}, \lambda)$$ is a hidden vector $\bm{z}$ and a sequence of \textit{ground truth} event types and event times. It is tasked with predicting the next type of event $\hat{k}$, the next event time $\hat{t}$, and the intensity function $\lambda$ that measures the probability of an event occurring. $\lambda$ is necessary to compute the likelihood $P_\theta((t_1, k_1), \dots, (t_i, k_i),|\bm{z})$.  
\footnote{Please see Appendix for details.}
Furthermore, we follow Transformer Hawkes Process's approach of also adding mean squared error losses to the time: $time\_loss = \|t - \hat{t}\|^2$ and cross-entropy loss of the predicted $type\_loss = -\sum_{c=1}^{|\mathcal{K}|} k \log(p_{k})$

At \textit{inference time}, the input to the decoder is only $z$, and we decode the predicted event types and times.
To predict next time and event tuple $(\hat{t}_i, \hat{k}_i)$, the input is the previously predicted times and events $\{(\hat{t}_1, \hat{k}_1), \dots, (\hat{t}_{i-1}, \hat{k}_{i-1}))\}$). (each predicted time and event is repeatedly appended to the input). 

Finally, we note that we can control for the generation of events that are similar to the original patient by first encoding the original patient and then sampling around it, a benefit of the probabilistic nature of the VAE latent space $z$. Otherwise, it would be impossible to correspond the original labels to the synthetic data. For all experiments in this work, we take a random sample of the latent vector $\bm{z}$ to reconstruct our patient. Otherwise, our task would collapse down to a straightforward autoencoder task.

\subsection{Final Loss Terms}

Finally, we write the final loss as $$L = L_{hawkes} + L_{elbo} + L_{length}$$
The $L_{hawkes}$ is the log-likelihood of the sequence given the Hawkes process above. The $L_{elbo}$ is the VAE loss of the hidden vector KL divergence from a standard Gaussian, the mean-squared error reconstruction loss of the event times, and the cross-entropy loss of the event types. Finally, we additionally add $L_{length}$ to ensure the model learns proper sequence lengths (described in section \ref{sec:length}).

\begin{figure}[ht]
    \centering
    \includegraphics[width=.8\linewidth, trim={0 .2cm 0 71.2cm},clip]{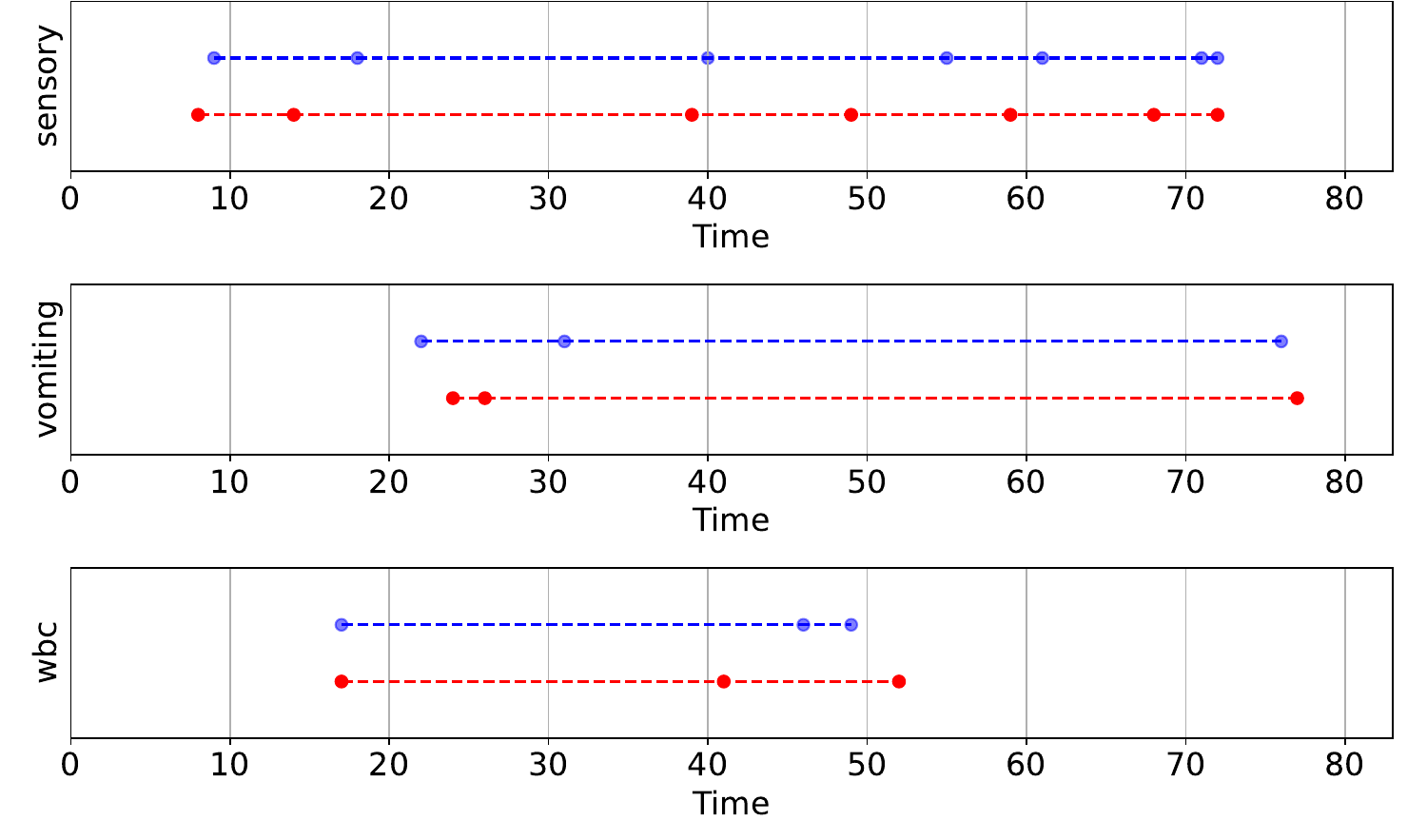}
    \caption{Example of a generated sequence from \method from NCT00003299 plotted by the individual events. Red dots and lines denote ground truth event occurrence and time between events respectively. In this case, the time is in Days. The blue dots and lines are the predicted events. Numerical events such as wbc (white blood cell count) are discretized based on their unique values in the real data. This will be corrected in the new version. Each prediction is linked with dashed lines for clarity.} 
    \label{fig:example}
    \vspace{-1em}
\end{figure}

\textbf{Numerical Values}
Note that we do not discretize the time in terms of the time gap. Rather, we pad out each event sequence to the number of the most occurrences, which is usually around 100-200. Each event is considered to be categorical, and numerical events such as wbc (white blood cell count in Figure~\ref{fig:example}) is discretized based on their unique values in real-world data. 

\begin{table*}[ht]

\caption{A description of all the real-world datasets used in the evaluation. All trial data was obtained from Project Data Sphere \citep{green2015project}. \textit{Num Rows} refers to the raw number of data points in the trial. 
\textit{Num Subj} refers to the total number of patients. \textit{Num Events} denotes the total number of \textit{unique} events. \textit{Events / Subj} denotes the average number of events that a patient experiences. \textit{Positive Label Proportion} denotes the percentage of patients that did not experience the death event.}
\label{tab:data}

\centering
\resizebox{\linewidth}{!}{  
\begin{tabular}{ccccccc}

Dataset &
  Description &
  \# Rows &
  \# Subjects &
  \# Events &
  Events / Subject &
  \begin{tabular}[c]{@{}c@{}}Positive Label \\ Proportion\end{tabular} \\ \midrule
NCT00003299 (LC1)& \begin{tabular}[c]{@{}c@{}}Small Cell \\ Lung Cancer \end{tabular} & 20210 & 548 & 34 & 36.880 & 0.951 \\ \midrule
NCT00041119 (BC1) & Breast Cancer                                                      & 2983  & 425 & 150 & 7.019  & 0.134 \\ \midrule
NCT00079274 (CC) & Colon Cancer                                                 & 316   & 70  & 18  & 4.514  & 0.184 \\ \midrule
NCT00174655 (BC2)     & Breast Cancer                                               & 7002  & 953 & 21  & 7.347  & 0.019 \\ \midrule
NCT00312208 (BC3) & Breast Cancer                                                      & 2193  & 378 & 182 & 5.802  & 0.184 \\ \midrule
NCT00694382 (VTE) &
  \begin{tabular}[c]{@{}c@{}}Venous \\ Thromboembolism \\ in Cancer Patients \end{tabular} &
  7853 &
  803 &
  746 &
  9.780 & 
  0.456 \\ \midrule
NCT03041311 (LC2) & \begin{tabular}[c]{@{}c@{}}Small Cell \\ Lung Cancer \end{tabular} & 1043  & 47  & 207 & 22.192 & 0.622\\
\bottomrule
\end{tabular}
}
\end{table*}

\subsection{Event Type Information}
We also propose 2 variants of \method. In some applications, such as clinical trial patient modeling~\citep{wang2022promptehr,fu2022hint,chen2024uncertainty,lu2024uncertainty,lu2019integrated,das2023twin}, we may be interested in an event sequence \textit{with known event types}, that is, the model only needs to generate the timestamps at which events occur. This is to address the concern of subject fidelity, that is, the generated subject must be significantly similar to the original subject in order for the generated data to be useful; therefore, knowing which events occur in a subject to generate a similar subject would not be unreasonable. Along with the ``Events Unknown'' model that has no assumptions, we also propose the ``Events Known'' model was created to enforce ONLY simulating specific events, without considering all events (which may be too numerous and irrelevant to the current patient). 

To accommodate \method (Events Known), we use the exact same model as \method (Events Unknown), but restrict the event type prediction module to only valid patient input event types at \textit{inference time}. We retain the same training process for both models, since we do not want to restrict learning event type information at training.


\subsection{Sequence Length Prediction}
\label{sec:length}

We generate event sequences $\{(t_j, k_j); j=1,\dots,L ; k_j \in \mathcal{K}'\}, $
where length $L$ is also generated by \method. 
Taking inspiration from HALO \cite{theodorou2023synthesize}, our generation process automatically appends an \texttt{[END]} event at the end of each of the patient events. Furthermore, in addition to the event loss from before, we add a cross-entropy loss term on specifically the [END] event. 


\section{Experiments}
\label{sec:Experiments}

\subsection{Datasets} 
We evaluated our models on 7 real-world clinical trial outcome datasets obtained from Project Data Sphere\footnote{\url{https://data.projectdatasphere.org/projectdatasphere/html/access}} \citep{green2015project,fu2023automated,lu2019integrated,chen2024uncertainty,lu2024uncertainty}. Specifically, we chose the trials as outlined in Table~\ref{tab:data}. These datasets have shown to be effective evaluation datasets for tabular prediction \citep{wang2022transtab, wang2023anypredict} and digital twin generation \citep{das2023twin,wang2024twin}. Specifically, we use LC1 \citep{niell2005randomized}, BC1 \citep{baldwin2012genome}, CC \citep{alberts2012effect}, BC2 \citep{fernandez2012prognostic},  BC3 \citep{wu20241596}, VTE \citep{agnelli2012semuloparin}, LC2 \citep{daniel2021trilaciclib,fu2024ddn3,zhang2021ddn2}. A full description of the data is shown in Table~\ref{tab:data}. 
Each dataset contains events and the times at which they occur, e.g., medications and procedures, as well as some adverse events like vomiting etc. We use these datasets to predict if the subject experiences the death event, which is an external label. Note that \method does not require a fixed patient event sequence length. 

\subsection{Baseline Methods} One surprising challenge we found was that \textit{existing EHR methods and synthetic patient generation methods are not applicable to our specific task and dataset due to dataset size and lack of support for timestamp generation}; therefore, we primarily compare against general sequential data generation methods.

We compared the following 7 models: First, the \textbf{LSTM VAE} is the same as our proposed model, except with an LSTM instead of a Transformer encoder. \textbf{PARSynthesizer} from is SDV, based on a conditional probabilistic auto-regressive (CPAR) model, is specifically tailored for synthesizing sequential event data and stands out due to its unique focus and accessible codebase. \textbf{TabDDPM} is a state-of-the-art tabular synthesizer using diffusion models, enhanced by adding time as a numerical column for our purposes. Despite not being explicitly designed for sequential data, it surpasses previous models like \textbf{CTGAN} in synthetic tabular data generation. Lastly, \textbf{HALO}, a hierarchical autoregressive language model, excels in synthesizing Electronic Health Records (EHR) but struggles with clinical trial datasets due to the limited size of the training data, highlighting the challenges in this domain.

\textbf{\method (Events Unknown)} is the VAE + Multivariate Hawkes Process that is trained without any assumptions. At training time, the task is to predict a patient's events and timesteps given the latent vector. 
\textbf{\method (Events Known)} assumes that one knows which specific events occur for the Hawkes Model. This essentially just restricts the number of valid events in the prediction phase by patient's unique events.

\subsection{Utility Evaluation}

\begin{table}[ht]
\centering
\caption{Utility Evaluation: Binary Classification ROCAUCs ($\uparrow$ higher the better, $\pm$ standard deviation) of a downstream LSTM trained on data generated from the \method models as well as the original data and baselines. Note that the LSTM and the \method models estimate their own sequence length. \method (Events Known) is put in a separate category due to its requirement of event type information, with results \textbf{underlined}. \textbf{Bolded} indicates original data ROC is within 1 standard deviation of synthetic data ROC}
\label{tab:main_results}
\setlength\tabcolsep{4pt} 
\resizebox{\linewidth}{!}{
\begin{tabular}{ccccccccc} 
Dataset &
  \begin{tabular}[c]{@{}c@{}}Original \\ Data\end{tabular} &
  \begin{tabular}[c]{@{}c@{}}LSTM \\VAE\end{tabular} &
  PAR  &CTGAN&
  TabDDPM  &HALO&
  \begin{tabular}[c]{@{}c@{}}\method\end{tabular} &  
  \begin{tabular}[c]{@{}c@{}}\method \\ (Events \\Known)\end{tabular} \\ \midrule 
LC1 & 0.689$_{\pm 0.105}$ & 0.563$_{\pm 0.053}$ & 0.504$_{\pm 0.066}$  &0.508$_{\pm 0.122}$& 0.557$_{\pm 0.055}$  &0.457$_{\pm 0.079}$& \textbf{0.672$_{\pm 0.061}$}& \textbf{0.709$_{\pm 0.049}$} \\
BC1 & 0.678$_{\pm 0.078}$ & 0.617$_{\pm 0.036}$ & 0.573$_{\pm 0.043}$  &0.550$_{\pm 0.046}$& \textbf{0.630$_{\pm 0.045}$}  &0.461$_{\pm 0.184}$& \textbf{0.651$_{\pm 0.046}$}& \textbf{0.665$_{\pm 0.045}$} \\
CC & 0.657$_{\pm 0.140}$& 0.481$_{\pm 0.092}$ &0.567$_{\pm 0.096}$&0.448$_{\pm 0.023}$& \textbf{0.583$_{\pm 0.098}$}&0.446$_{\pm 0.02}$&\textbf{ 0.652$_{\pm 0.015}$}& \textbf{0.653$_{\pm 0.019}$} \\
BC2 & 0.660$_{\pm 0.128}$ & 0.535$_{\pm 0.073}$ & 0.523$_{\pm 0.074}$  &0.523$_{\pm 0.11}$& 0.513$_{\pm 0.078}$  &0.503$_{\pm 0.075}$& \textbf{0.599$_{\pm 0.042}$}& \textbf{0.594$_{\pm 0.068}$} \\
BC3 & 0.632$_{\pm 0.072}$ & 0.454$_{\pm 0.039}$ & 0.463$_{\pm 0.039}$  &0.493$_{\pm 0.013}$& 0.503$_{\pm 0.043}$  &0.535$_{\pm 0.183}$& \textbf{0.620$_{\pm 0.038}$}& \textbf{0.634$_{\pm 0.032}$} \\
VTE & 0.640$_{\pm 0.038}$ & 0.490$_{\pm 0.019}$ & 0.549$_{\pm 0.022}$  &0.508$_{\pm 0.113}$& 0.531$_{\pm 0.021}$  &0.485$_{\pm 0.066}$& \textbf{0.618$_{\pm 0.024}$}& \textbf{0.625$_{\pm 0.020}$} \\
LC2 & 0.738$_{\pm 0.149}$ & 0.563$_{\pm 0.097}$ & 0.507$_{\pm 0.087}$  &0.573$_{\pm 0.118}$& 0.574$_{\pm 0.096}$  &0.534$_{\pm 0.078}$& \textbf{0.729$_{\pm 0.044}$}& \textbf{0.755$_{\pm 0.059}$} \\ \bottomrule 
\end{tabular}
}
\end{table}

\textbf{Downstream Classification ROCAUC:} It is vital that synthetic data perform similarly to real-world data; therefore, we evaluate the utility (ROCAUC) of the generated synthetic data by performing binary classification of death events in all 7 clinical trials. We choose ROCAUC since it has been used for similar tasks in the past\cite{das2023twin}. Additionally, ROC AUC is sensitive to class imbalance in the sense that when there is a minority class, one typically defines this as the positive class and it will have a strong impact on the AUC value. This is desirable behavior and is what we look to evaluate in our application.

The standard deviation of each ROCAUC score is calculated via bootstrapping (100x bootstrapped test data points). 
Training is performed completely on synthetic data by matching each generated patient to its ground truth death event label. Testing is performed on the original held-out ground truth split. For the Original Data baseline, we performed 5 cross-validations on 80/20 train test splits of the real data. The main results are shown in Table~\ref{tab:main_results}. 

We see that synthetic data generated by \method variants generally perform the best in terms of downstream death event classification performance, where \method (Events Unknown) outperforms the next best model (in 4/7 datasets and is within 1 standard deviation with the rest of the datasets). Furthermore, \method (Events Known) significantly outperforms other baselines, due to the additional input information. Still, \method (Events Unknown) also performs admirably, being on par but slightly less performant than \method (Events Known).

Occasionally, synthetic data is able to support better performance than the original dataset on downstream tasks (this behavior is also seen in TabDDPM). We believe that this is due to the synthetic model generating examples that are more easily separable and/or more diverse than real data. However, this is only a hypothesis and should be investigated further in future research, but we are encouraged to see that our proposed method captures this interesting synthetic data behavior.

\begin{table}[ht]
\caption{Results of ML Inference Score: LSTM binary classification of real vs synthetic (\textit{the closer to 0.5 the score is, the better}). The standard deviation calculated via bootstrapping is shown via $\pm$. AUCROC scores are shown. Bolded indicates the best result or within 1 standard deviation of the best result. }
\label{tab:privacy}
\centering
\resizebox{\linewidth}{!}{  
\begin{tabular}{cccccccc}
\toprule
Dataset & \begin{tabular}[c]{@{}c@{}}LSTM \\VAE\end{tabular} & PAR  &CTGAN& TabDDPM  &HALO& 
\begin{tabular}[c]{@{}c@{}}\method\end{tabular}  
&\begin{tabular}[c]{@{}c@{}}\method\\  (Events \\Known)\end{tabular} \\ \midrule
LC1 & 1.000$_{\pm 0.000}$ & 0.968$_{\pm 0.010}$  &0.952$_{\pm 0.056}$& 0.762$_{\pm 0.024}$  &1.000$_{\pm 0.004}$& \textbf{0.613$_{\pm 0.024}$}& 0.689$_{\pm 0.020}$\\
BC1 & 0.932$_{\pm 0.017}$ & 0.998$_{\pm 0.002}$  &0.973$_{\pm 0.082}$& 0.926$_{\pm 0.017}$  &1.000$_{\pm 0.001}$& \textbf{0.616$_{\pm 0.025}$}& 0.768$_{\pm 0.021}$ \\
CC & 1.000$_{\pm 0.000}$ & 0.807$_{\pm 0.082}$  &0.935$_{\pm 0.056}$& 0.894$_{\pm 0.050}$  &0.998$_{\pm 0.005}$& 0.711$_{\pm 0.051}$& \textbf{0.701$_{\pm 0.054}$} \\
BC2 & 1.000$_{\pm 0.000}$ & 0.999$_{\pm 0.001}$  &0.998$_{\pm 0.075}$& 0.998$_{\pm 0.001}$  &0.999$_{\pm 0.001}$& \textbf{0.605$_{\pm 0.048}$}& \textbf{0.593$_{\pm 0.023}$} \\
BC3 & 0.994$_{\pm 0.007}$ & 0.874$_{\pm 0.026}$  &0.895$_{\pm 0.098}$& 0.729$_{\pm 0.035}$  &0.992$_{\pm 0.008}$&\textbf{0.689$_{\pm 0.023}$}& \textbf{0.693$_{\pm 0.038}$} \\
VTE & 1.000$_{\pm 0.000}$ & 0.923$_{\pm 0.012}$  &0.879$_{\pm 0.119}$& 0.992$_{\pm 0.005}$  &0.000$_{\pm 0.004}$& \textbf{0.871$_{\pm 0.014}$}& \textbf{0.856$_{\pm 0.016}$} \\
LC2 & 1.000$_{\pm 0.000}$ & 0.651$_{\pm 0.112}$  &0.982$_{\pm 0.038}$& 0.374$_{\pm 0.021}$  &0.000$_{\pm 0.003}$& \textbf{0.573$_{\pm 0.111}$}  & \textbf{0.477$_{\pm 0.127}$} \\ \bottomrule
\end{tabular}
}
\end{table}
\subsection{Privacy evaluations}

\textbf{ML Inference Score}: This can also be thought of as an adversarial Model Attack \cite{theodorou2023synthesize}. Another main concern is the privacy of the synthetic data, to prevent any data or information leakage. To address this, we calculate the performance of predicting whether a generated sequence is real vs synthetic via an LSTM binary classification \citep{SDV} (similar to an adversarial model). 
The real subjects are labeled with ``0'' and the synthetic subjects are labelled with ``1''. Results are shown in Table~\ref{tab:privacy}, and we see that \method variants perform closest to the optimal 0.5 ROCAUC ideal score. One thing to note is that a perfect copy of the original data would result in a 0.5 score, so we have the following metric to measure the opposite scenario. Furthermore, we see a continued trend of both forms of \method generally outperforming other baseline methods, illustrating the importance of giving the model more information in this data-scarce setting.


\begin{wraptable}{r}{.58\linewidth}
\vspace{-1.9em}

\caption{Distance to Closest Record (DCR) Score. Note that this score only tells part of the picture. The higher this score is, the larger the difference between the synthetic data and the original data. The lower the score, the more similar the synthetic data is to the original data.}
\label{tab:privacy2}

\centering
\resizebox{\linewidth}{!}{
\begin{tabular}{cccccc} 
Dataset & \begin{tabular}[c]{@{}c@{}}LSTM \\VAE\end{tabular} & PAR & TabDDPM & \begin{tabular}[c]{@{}c@{}}\method \\ (Events \\ Unknown)\end{tabular}  &\begin{tabular}[c]{@{}c@{}}\method \\ (Events \\ Known)\end{tabular} \\ \midrule
LC1 & 3.700 & 2.647 & 1.426 & 1.217& 1.138 \\
BC1 & 4.677 & 4.633 & 1.007 & 0.624& 0.612 \\
CC & 2.732 & 1.977 & 1.346 & 1.519& 1.675 \\
BC2 & 32.185 & 56.915 & 3.581 & 1.452& 1.215  \\
BC3 & 87.015 & 2.348  & 1.207 & 0.515& 0.745 \\
VTE & 17.946 & 35.362  & 1.059 & 0.983& 0.971 \\
LC2 & 36.740 & 37.723 & 4.662 & 5.015& 4.922 \\ \bottomrule
\end{tabular}
}

\caption{Dataset Inference attack: \textit{(the closer to .5 the better)}. This is calculated as the percent where the closest record of a training sample is a real vs synthetic sample.}
\label{tab:datasetattack}

\resizebox{\linewidth}{!}{
\begin{tabular}{cccccccc} 
Dataset & \begin{tabular}[c]{@{}c@{}} {\small LSTM} \\{\small VAE}\end{tabular} & {\small PAR} & {\small CTGAN} & {\small HALO} & {\small TabDDPM} & \begin{tabular}[c]{@{}c@{}} {\small TrialSynth }\\ {\small (Events} \\ {\small Unknown)}\end{tabular} & \begin{tabular}[c]{@{}c@{}} {\small TrialSynth} \\ {\small (Events} \\ {\small Known)} \end{tabular} \\ \midrule 
LC1 & 1.00 & 0.99 & 0.97 & 0.98 & 0.71 & 0.62& 0.59\\
BC1 & 1.00 & 0.92 & 0.84 & 1.00 & 0.71 & 0.61& 0.52 \\
CC & 0.97 & 0.87 & 0.81 & 1.00 & 0.42 & 0.62& 0.38 \\
BC2 & 1.00 & 0.98 & 0.97 & 0.99 & 0.99 & 0.73 & 0.62 \\
BC3 & 0.99 & 0.77 & 0.60 & 1.00 & 0.35 & 0.40& 0.44 \\
VTE & 1.00 & 0.89 & 0.65 & 1.00 & 0.86 & 0.87& 0.37 \\
LC2 & 1.00 & 1.00 & 0.91 & 1.00 & 0.27 & 0.62& 0.25 \\ \bottomrule
\end{tabular}
}
\vspace{-1em}
\end{wraptable}

\textbf{Distance to Closest Record (DCR) Score}: Second, we follow the evaluation metrics per TabDDPM \citep{kotelnikov2023tabddpm}. That is,  we compare the feature vectors of the real vs synthetic data and measure how far the synthetic data is from the original. The higher this distance is, the more different the generated data is from the original data, and thus the more private it is. A completely different version of the data would obtain the highest distance but could result in bad performance in the downstream LSTM classification performance or a high ML Inference score (close to 1). We calculate this by featurizing the event time predictions in terms of (count, means, and standard deviations). Then, we normalize and obtain the L2 distance between a generated subject and the closest real subject. Table \ref{tab:privacy2} shows this result. 
Notice that \method variants generally obtain quite low scores on this metric. TabDDPM and PAR also generate data closer to the original data compared to LSTM VAE. We note the privacy-fidelity trade-off, as LSTM VAE generates data that is further away from the original, but yields worse utility (Table~\ref{tab:main_results}).

\textbf{Dataset Attack} We evaluate a Dataset Attack scenario as per HALO \cite{theodorou2023synthesize}, where we label the real records with the lowest distance (computed by featuring event times into mean, std, counts) to the closest record in the synthetic dataset as 1. It tests the ability of the synthetic dataset to prevent an attacker from inferring whether a real record was used in the training dataset. On real training data, we compare if the closest record is a real record from the training set or a synthetic record. Ideally, we also want this accuracy to be 0.5. 
From Table~\ref{tab:datasetattack}, we see that \method generally performs the best, even beating out HALO and TabDDPM.

\subsection{Utility / Privacy Trade-off}

\begin{figure}[ht]
    \centering
    \includegraphics[width=0.49\linewidth]{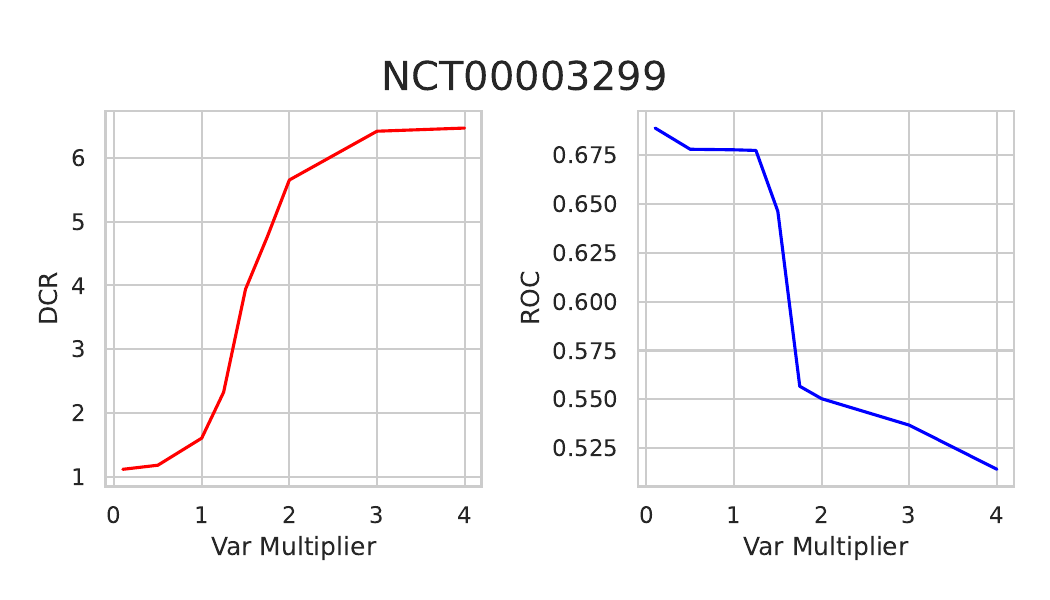}
    \includegraphics[width=0.49\linewidth]{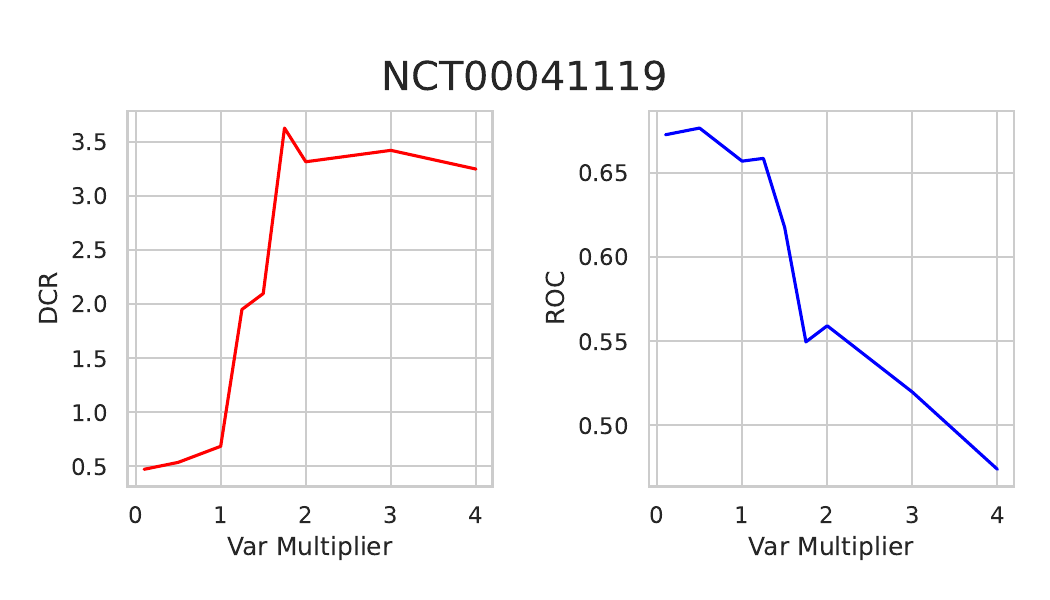}
    \caption{2 Privacy-Utility Tradeoff examples in \method: Performance of distance to closest record (DCR) (red) and downstream ROC (blue) metrics at varying levels of VAE sampling variance (from 0.1 to 4), represented as the ``Var Multiplier.''}
    \label{fig:utility_privacy}
\end{figure}

In \method, the privacy-utility tradeoff is governed by the variance applied to the VAE sampling process (Figure~\ref{fig:utility_privacy} and Figure~\ref{fig:add_ablation}). Increasing the variance in VAE sampling introduces more diversity into the synthetic data, enhancing privacy by making it harder to trace back to original data points. However, as the Var Multiplier rises, the quality of utility metrics such as downstream ROC tends to decrease, reflecting a drop in predictive accuracy and utility for downstream tasks. Conversely, metrics like DCR may rise, indicating a more extensive departure from the original dataset.
A unique advantage of \method is its capacity to provide direct control over the tradeoff between fidelity and privacy through the adjustment of VAE sampling variance. By tuning this "Var Multiplier," researchers can precisely regulate how closely the synthetic data resembles the original dataset. Lower variance settings yield data with higher fidelity, making it more useful for predictive analyses and downstream clinical tasks, while higher variance introduces greater diversity, enhancing privacy protections by reducing the likelihood of re-identifying individual patients.

\section{Discussion}
\label{sec:Discussion}

The study presents \method, an innovative model that combines Variational Autoencoders (VAE) with Hawkes Processes (HP) to generate realistic synthetic sequential clinical trial data. Designed to address the challenges of small patient populations and the need for detailed time-event sequences, \method effectively captures both the timing and type of clinical events with high fidelity. Compared to existing methods, it outperforms in preserving data utility for downstream tasks while maintaining robust privacy protections, making it difficult to distinguish synthetic data from real data. 
Specifically, we demonstrate that \method outperforms existing methods in terms of data utility, enabling the generation of highly authentic event sequences across multiple real-world sequential event datasets. Empirical experiments indicate that providing the model with additional information, such as event index (Events Known) or event length, leads to significant improvements in the synthetic data quality. Finally, we believe that a sweet spot is reached by allowing the model to know the event index--as it provides a significant downstream classification boost while maintaining a low ML inference score, and is a common assumption when generating specific patients. We note that relaxing this assumption still yields competitive performance.
Overall, \method offers a powerful solution for synthetic data generation in healthcare, balancing patient privacy with data authenticity, and shows promise for broader applications in clinical trial design and other healthcare domains that demand high-quality, secure synthetic datasets.


\bibliography{bib}
\bibliographystyle{plain}

\clearpage
\DoToC
\appendix
\section{Appendix}

\subsection{Limitations}
The paper presents a promising method for generating synthetic time-sequential clinical trial data, but there are several limitations to consider. First, the generalizability of \method may be restricted, as its performance is demonstrated on small patient populations, leaving its effectiveness on larger, more diverse datasets uncertain. Additionally, while the use of Hawkes Processes (HP) helps model event-type and time gap prediction, this approach may struggle with more complex or non-linear temporal dynamics seen in real-world clinical data. Another limitation lies in the interpretability of the model. As a Variational Autoencoder (VAE), \method can be challenging to interpret compared to more traditional models, which is a crucial aspect when applying the method to clinical scenarios.

While the paper asserts that \method empirically preserves patient privacy, it lacks a comprehensive assessment of potential re-identification risks, leaving questions about the robustness of its privacy-preserving capabilities. Moreover, while the utility of the generated data is demonstrated in specific contexts, the broader applicability of the synthetic data, such as in clinical trials or regulatory processes, remains underexplored, (but this is a problem endemic to the field as a whole.) 

\subsection{Societal Impact}
The societal impact of the proposed method for generating synthetic time-sequential clinical trial data has several promising positive aspects, with a few notable challenges. On the positive side, the ability to generate high-fidelity synthetic clinical data can significantly accelerate the pace of medical research and the development of new treatments. By simulating patient trajectories, researchers can optimize trial designs, potentially reducing the time and cost required to bring life-saving interventions to market. This could lead to faster availability of new drugs and treatments, especially for rare diseases or conditions where patient recruitment for trials is challenging. Additionally, synthetic data can alleviate privacy concerns, as it reduces the reliance on real patient data, thereby protecting sensitive personal information while still enabling valuable research. This would empower institutions to collaborate and share data more freely, further advancing innovation.

Another significant societal benefit lies in improving equity in healthcare research. Many populations are underrepresented in clinical trials due to geographic, socio-economic, or logistical barriers. Synthetic data generation can help address this imbalance by allowing researchers to simulate the effects of treatments on diverse populations, leading to more inclusive healthcare solutions. This could help mitigate health disparities by ensuring new treatments are designed with a broader range of patient needs in mind.

However, there are some societal challenges to consider. One potential negative impact is the over-reliance on synthetic data, which, despite its fidelity, is not a perfect substitute for real-world clinical data. There is a risk that inaccuracies in the synthetic data could lead to suboptimal clinical decisions if the limitations are not adequately understood. Additionally, while synthetic data can protect patient privacy, concerns about data security and the potential for misuse of generated data still remain. Mismanagement of synthetic data could undermine trust in medical research, particularly if stakeholders perceive it as less reliable than traditional methods.

\subsection{\method Details}

\textbf{Neural Hawkes Processes} are formulated as follows.
We are given a set of $L$ observations of the form (time $t_j$, event\_type $k_j$). $S = \{(t_1, k_1), \dots, (t_j, k_j), \dots, (t_L, k_L)\}$
Each time $t_j \in \mathbb{R}^+ \bigcup \{0\}$ and is sorted such that $t_j < t_{j+1}$. Each event $k_j \in \{1, \dots, K\}$.
The traditional Hawkes Process assumption that events only have a positive, decaying influence on future events is not realistic in practice, as there exist examples where an occurrence of an event lowers the probability of a future event (e.g., medication reduces the probability of adverse events). Therefore, the Neural Hawkes Process \citep{mei2017neural} was proposed to generalize the traditional Hawkes Process. The following derivations follow \citep{zuo2020transformer}.

$$\lambda(t) := \sum_{k=1}^K \lambda_k(t) := \sum_{k=1}^K f_k(\bm{W}_k^\top \bm{h}(t)) = \sum_{k=1}^K \beta_k \log \left(1+ e^{\frac{\bm{W}_k^T \bm{h}(t)}{\beta_k}} \right), $$
where $\lambda(t)$ is the intensity function for \textit{any} event occurring, $\lambda_k(t)$ is the intensity function for the event $k \in \mathcal{K}$ occurring, $K = |\mathcal{K}|$ is the total number of event types, $h(t)$ are the hidden states of the event sequence obtained by a Transformer encoder, and $\bm{W}_k^\top$ are learned weights that calculate the significance of each event type at time $t$. 

$f_k(c) = \beta_k \log(1+e^{\frac{x}{\beta_k}})$ is the softplus function with parameter $\beta_k$. The output of $f_k(x)$ is always positive. Note that the positive intensity does not mean that the influence is always positive, as the influence of previous events are calculated through $\bm{W}_k^\top \bm{h}(t)$. If there is an event occurring at time $t$, then the probability of event $k$ is $P(k_t=k) = \frac{\lambda_k(t)}{\lambda(t)}$.

Let the history of all events before $t$ be represented by $\mathcal{H}_t = \{(t_j, k_j), t_j<t\}$. The continuous time intensity for prediction is defined as
$$\lambda(t | \mathcal{H}_{t}) := \sum_{k=1}^K \lambda_k(t | \mathcal{H}_{t}) := \sum_{k=1}^K f_k \left( \alpha_k \frac{t-t_j}{t_j} + \bm{W}_k^\top \bm{h}(t_j) + \mu_k \right),$$
where time is defined on interval $[t_j, t_{j+1})$, $f_k$ is the softplus function as before, $\alpha_k$ is a learned importance of the interpolation between the two observed timesteps $t_j$ and $t_{j+1}$.
Note that when $t=t_j$, $\alpha_k$ does not matter as the influence is $0$ (intuitively, this is because we know that this event exists, so there is no need to estimate anything). The history of all previous events up to time $t$ is represented by $\bm{t}_j$. $\bm{W}^\top_k$ are weights that convert this history to a scalar. $\mu_k$ is the base intensity of event $k$.
Therefore, the probability of $p(t|\mathcal{H}_{t_j})$ is the intensity at $t \in [t_j, t_{j+1})$ given the history $\mathcal{H}_t$ and the probability that no other events occur from the interval $(t_j, t)$
$$p(t|\mathcal{H}_{t_j}) = \lambda(t|\mathcal{H}_{t}) \exp{\left( -\int_{t_j}^t \lambda(t'|\mathcal{H}_{t'}) dt' \right)}. $$

Note that if $t_j$ is the last observation, then $t_{j+1}=\infty$.
Finally, the next time value $\hat{t}_{j+1}$ and event prediction $\hat{k}_{j+1}$ is given as 
$$\hat{t}_{j+1} = \int_{t_j}^\infty t \cdot p(t|\mathcal{H}_t) dt, \ \ \ \  
\hat{k}_{j+1} = \text{argmax}_{k}\frac{\lambda_k(t_{j+1}|\mathcal{H}_{t_{j+1}})}{\lambda(t_{j+1}|\mathcal{H}_{t_{j+1}})}
$$

For training, we want to maximize the likelihood of the observed sequence $\{(t_1, k_1), \dots, (t_L, k_L)\}$. The log-likelihood function is given by\footnote{The proof is shown in \citep{mei2017neural}}
$$\ell(\{(t_1, k_1), \dots, (t_L, k_L)\}) = \sum_{j=1}^L \log(\lambda(t_j| \mathcal{H}_{t_j})) - \int_{t_1}^{t_L} \lambda(t | \mathcal{H}_t) dt. $$

Finally, since the gradient of the log-likelihood function has an intractable integral, one may obtain an unbiased estimate by performing Monte Carlo sampling \citep{robert1999monte}.
$$\nabla \left[ \int_{t_1}^{t_L} \lambda(t | \mathcal{H}_t) dt \right]_{MC} = \sum_{j=2}^L (t_j - t_{j-1}) (\frac{1}{N} \sum_{i=1}^N \nabla\lambda(u_i))$$ With $u_i\sim Uniform(t_{j-1}, t_j)$. $\nabla\lambda(u_i)$ is fully differentiable with respect to $u_i$. 


Figure~\ref{fig:multiple} shows an example of the proposed model with all optional structural constraints (allowing the model to access the true event knowledge, such as type and event length information). 
To combine the VAE and the Hawkes process, we realize that the log-likelihood can be modeled as the log-likelihood of a Hawkes process if we assume that the event times $t$ and event types $k$ are generated from a Multinomial Gaussian, i.e., the combined loss may be written as the following.

Sample event sequence $S_z \sim P_\theta(S|z)$ where $$S_z = \{(t_1, k_1), \dots, (t_{L}, k_{L})\}$$ Then $H_{t,z}$ denotes the history up to time $t$ in $S_z$.

\begin{align} \notag
\lambda_\theta(t | \mathcal{H}_{t,z}) := \sum_{k=1}^K \lambda_{\theta,k}(t | \mathcal{H}_{t,z}) 
\notag = \sum_{k=1}^K f_k \left( \alpha_k \frac{t-t_{j}}{t_{j}} + \bm{W}_{\theta,k}^\top \bm{h}_\theta(t_{j}) + \mu_{\theta,k} \right)    
\end{align}
Where $t \in [t_{j}, t_{j+1})$. That is, $t$ lies between the $j$th and $j+1$th observation in $S_z$ (if $t_{j}$ is the last observation, then $t_{j+1}=\infty$).  $\lambda_{\theta,k}$,  $\bm{W}^\top_{\theta, k}$, and $\bm{h}^\top_{\theta}$ are the same as the Neural Hawkes process, only parameterized by $\theta$.

The log-likelihood is:
$$
\ln P_\theta(S_z|z) = \sum_{j=1}^{L} \log(\lambda_\theta(t_{j}| \mathcal{H}_{t_{j},z})) - \int_{t_1}^{t_{L}} \lambda_\theta(t | \mathcal{H}_{t,z})  dt. $$

For the VAE loss, we want to minimize the Kullback–Leibler divergence between $q_\phi(z|x)$ and $p_\theta(z|x)$, which in practice leads to maximizing the evidence lower bound (ELBO) for training along with the likelihood of $x$ \citep{kingma2013auto}.
$$L_{\theta, \phi} = \mathbb{E}_{z\sim q_\phi(\cdot|x)}[\ln P_\theta(x|z)] - D_{KL}(q_\phi(\cdot|x) || P_\theta(\cdot)). $$

Adding the VAE ELBO loss,  the combined \method loss is:
$$L_{\theta, \phi} = \mathbb{E}_{z\sim q_\phi(\cdot|S_z)}\left[ \ln P_\theta(S_z|z)
\right] - D_{KL}(q_\phi(\cdot|S_z) || P_\theta(\cdot| S_z) ). $$

\subsection{Ethics and Reproducibility}
Transformer Hawkes \citep{zuo2020transformer} is open source and can be found at \url{https://github.com/SimiaoZuo/Transformer-Hawkes-Process}. Training on an NVIDIA GeForce RTX 3090 takes around 12 hrs to run the full model. The code will be made public and open source on GitHub. for the camera-ready version.
All datasets were obtained from Project Data Sphere \citep{green2015project} with permission via a research data access request form. The links are as follows:
\begin{enumerate}
    \item NCT00003299 \citep{niell2005randomized}: A Randomized Phase III Study Comparing Etoposide and Cisplatin With Etoposide, Cisplatin and Paclitaxel in Patients With Extensive Small Cell Lung Cancer. Available at \url{https://data.projectdatasphere.org/projectdatasphere/html/content/261}
    
    \item NCT00041119 \citep{baldwin2012genome}: Cyclophosphamide And Doxorubicin (CA) (4 VS 6 Cycles) Versus Paclitaxel (4 VS 6 Cycles) As Adjuvant Therapy For Breast Cancer in Women With 0-3 Positive Axillary Lymph Nodes:A 2X2 Factorial Phase III Randomized Study. Available at \url{https://data.projectdatasphere.org/projectdatasphere/html/content/486}

    \item NCT00079274 \citep{alberts2012effect}: A Randomized Phase III Trial of Oxaliplatin (OXAL) Plus 5-Fluorouracil (5-FU)/Leucovorin (CF) With or Without Cetuximab (C225) After Curative Resection for Patients With Stage III Colon Cancer. Available at \url{https://data.projectdatasphere.org/projectdatasphere/html/content/407}
    
    \item NCT00174655 \citep{fernandez2012prognostic}: An Intergroup Phase III Trial to Evaluate the Activity of Docetaxel, Given Either Sequentially or in Combination With Doxorubicin, Followed by CMF, in Comparison to Doxorubicin Alone or in Combination With Cyclophosphamide, Followed by CMF, in the Adjuvant Treatment of Node-positive Breast Cancer Patients. Available at \url{https://data.projectdatasphere.org/projectdatasphere/html/content/127}
    
    \item NCT00312208 \citep{mackey2016long}: A Multicenter Phase III Randomized Trial Comparing Docetaxel in Combination With Doxorubicin and Cyclophosphamide Versus Doxorubicin and Cyclophosphamide Followed by Docetaxel as Adjuvant Treatment of Operable Breast Cancer HER2neu Negative Patients With Positive Axillary Lymph Nodes. Available at \url{https://data.projectdatasphere.org/projectdatasphere/html/content/118}
    
    \item NCT00694382 \citep{agnelli2012semuloparin}: A Multinational, Randomized, Double-Blind, Placebo-controlled Study to Evaluate the Efficacy and Safety of AVE5026 in the Prevention of Venous Thromboembolism (VTE) in Cancer Patients at High Risk for VTE and Who Are Undergoing Chemotherapy. Available at \url{https://data.projectdatasphere.org/projectdatasphere/html/content/119}
    
    \item NCT03041311 \citep{daniel2021trilaciclib}: Phase 2 Study of Carboplatin, Etoposide, and Atezolizumab With or Without Trilaciclib in Patients With Untreated Extensive-Stage Small Cell Lung Cancer (SCLC). Available at \url{https://data.projectdatasphere.org/projectdatasphere/html/content/435}
\end{enumerate}

\subsection{Baselines}
We describe our baselines in this section.

\textbf{LSTM VAE}: To compare against a VAE baseline, we manually implement our own LSTM VAE, which predicts the event type as a categorical classification task and the timestamp as a regression task at each event prediction.

\textbf{PARSynthesizer} from SDV \citep{par, SDV} since it is the most relevant model for synthesizing sequential event data, based on a conditional probabilistic auto-regressive (CPAR) model. To the best of our knowledge, no other models specifically handle sequential event data generation from scratch with easily accessible code.

    
\textbf{TabDDPM} \citep{kotelnikov2023tabddpm} is a recently proposed state-of-the-art general tabular synthesizer based on diffusion models. Although it is not explicitly built for sequential data, we are able to enhance it by adding time as a numerical column. This model also outperforms \textbf{CTGAN} models \cite{ctgan, zhao2021ctab, zhao2022ctab}, the previous go-to for synthetic tabular data generation. We believe that this is a strong, representative baseline of general tabular synthetic data generation. 

\textbf{HALO} \citep{theodorou2023synthesize} is state-of-the art hierarchical autoregressive language model that has achieved state-of-the-art performance for Electronic Health Record (EHR) synthesis. Still, it does not perform well on the clinical trial evaluation datasets, primarily due to the small size of training data, demonstrating the difficulty of this task.




\subsection{ML Utility Calculation Hyperparameters}
This section outlines hyperparameters explored for the downstream model for downstream ML Utility. 
\begin{table}[ht]
    \centering
    \caption{Hyperparameters Considered for LSTM Predictor Models}
    \label{tab:hyperparams3}
    \begin{tabular}{cc} 
    Parameter & Space \\ \midrule
    \texttt{embedding\_size} & [32,64,128] \\
    \texttt{num\_lstm\_layers} (Encoder) & [1,2] \\
    \texttt{hidden\_size} (Encoder) & [32,64,128] \\
    \texttt{lr} & [1e-3, 1e-4] \\
    \bottomrule
    \end{tabular}
\end{table}

\clearpage

\subsection{Examples}
\begin{figure*}[ht]
    \centering
    \includegraphics[width=.6\linewidth, trim={1.2cm 0cm .1cm 5cm},clip]{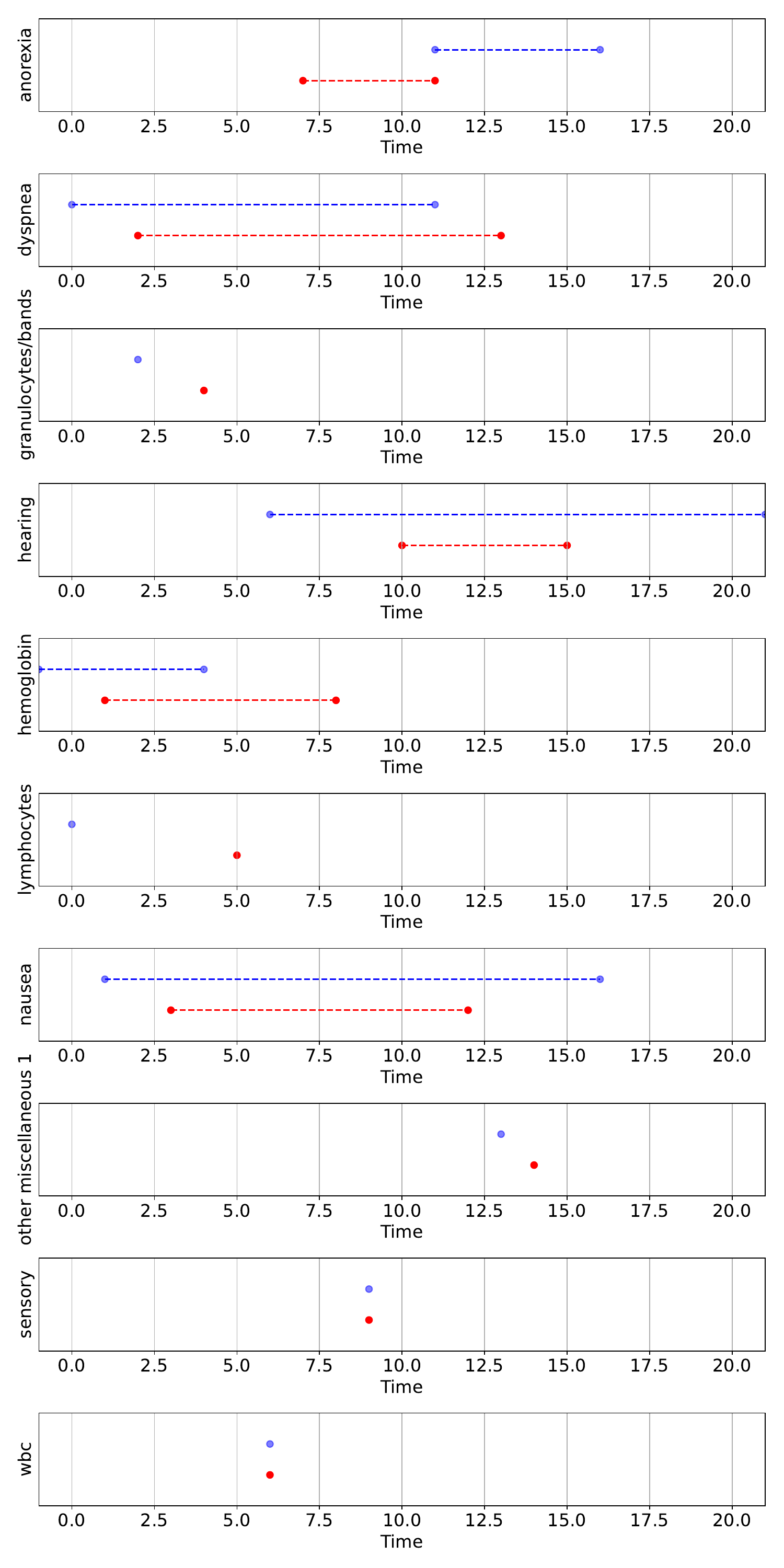}
    \caption{Example of another generated sequence from \method (Events Known) from NCT00003299. The \textcolor{blue}{blue dots} denoting the specific event timestamp prediction. The \textcolor{red}{red dots} are the ground truth timestamps and the ground truth predictions. Each prediction is also linked with dashed lines for clarity
    \label{fig:subjects1}}
\end{figure*}

\begin{figure*}[ht]
    \centering
    \includegraphics[width=.6\linewidth, trim={1.2cm 0cm .1cm 0cm},clip]{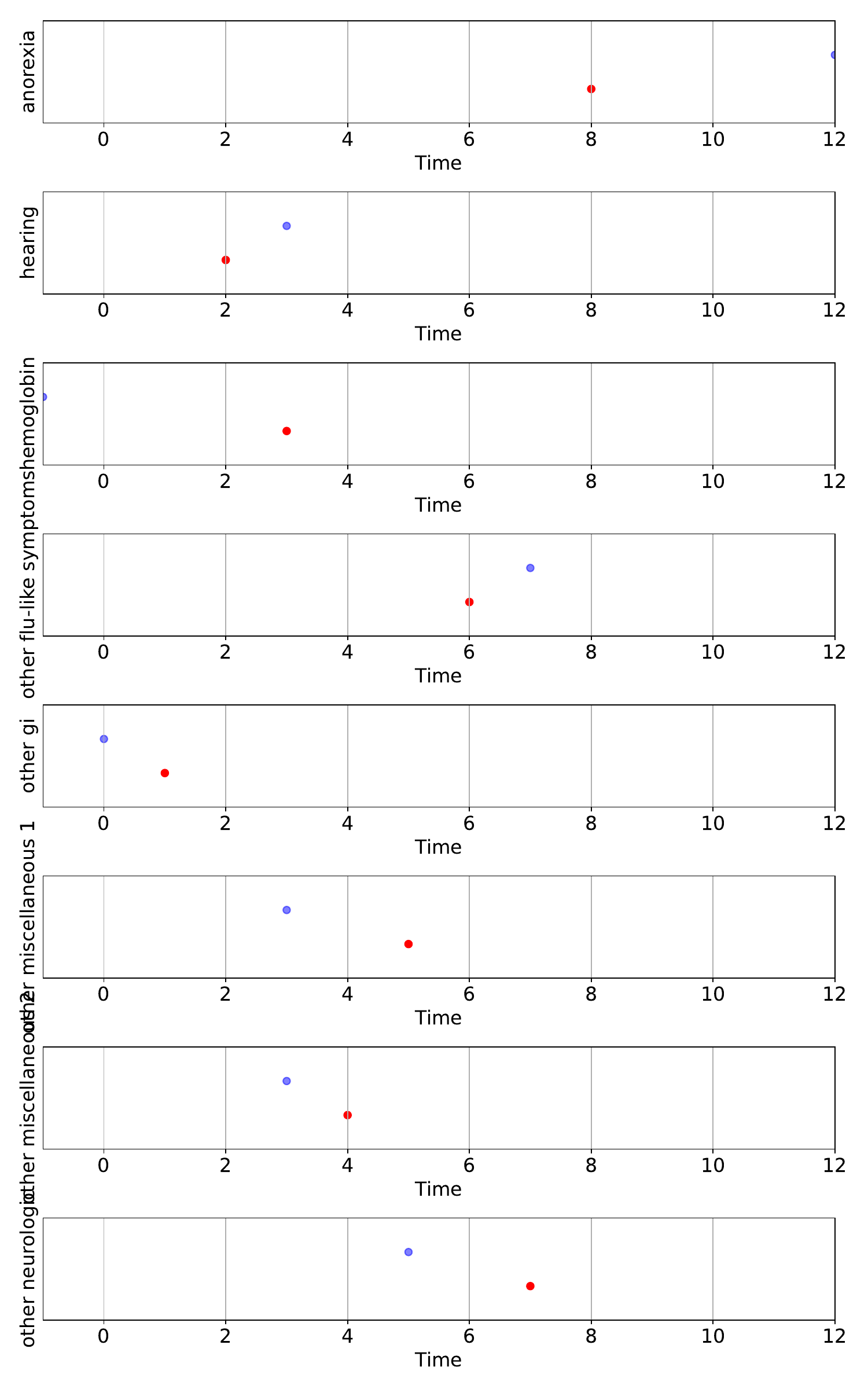}
    \caption{Example of another regenerated (encoded and decoded) sequence from \method (Events Known) from NCT00003299. The \textcolor{blue}{blue dots} denoting the specific event timestamp prediction. The \textcolor{red}{red dots} are the ground truth timestamps and the ground truth predictions. Each prediction is also linked with dashed lines for clarity \label{fig:subjects2}}
\end{figure*}

Figure~\ref{fig:subjects1} and Figure~\ref{fig:subjects2} show some examples of reconstructed subjects as generated by the best-performing model (\method (Events Known)). Intuitively, it visually reveals that the generated data generally matches the original data.

\clearpage
\subsection{Ablations}
\label{sec:ablations}
\begin{figure}[ht]
    \centering
    \includegraphics[width=0.49\linewidth]{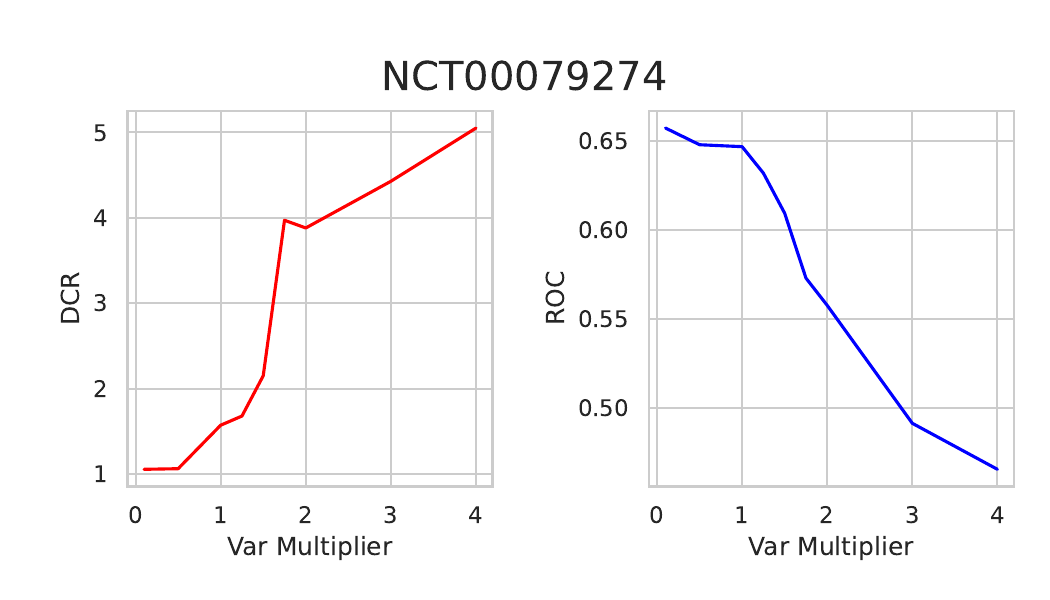}
    \includegraphics[width=0.49\linewidth]{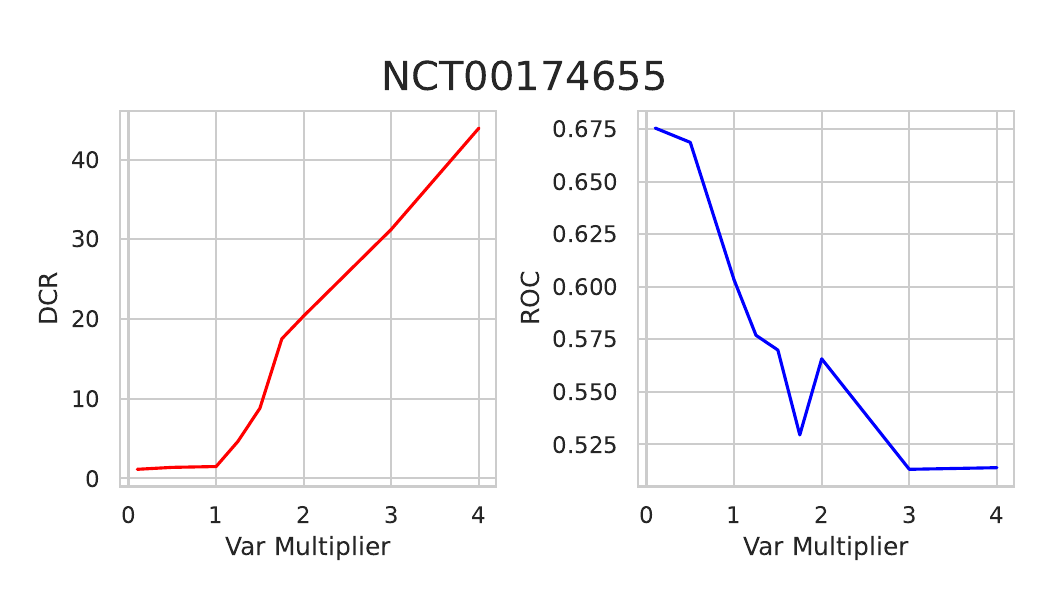}
    \includegraphics[width=0.49\linewidth]{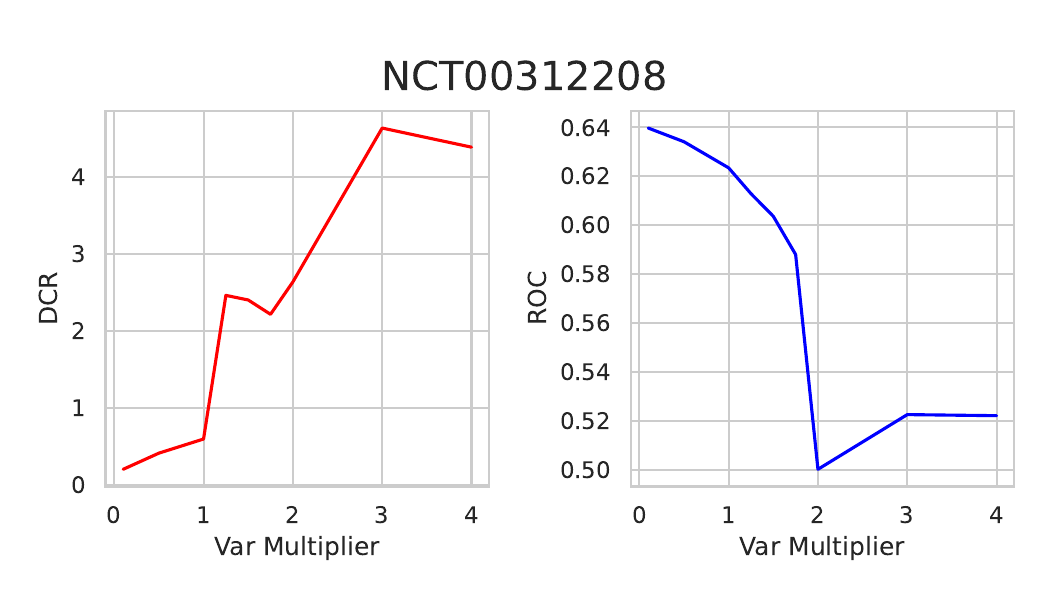}
    \includegraphics[width=0.49\linewidth]{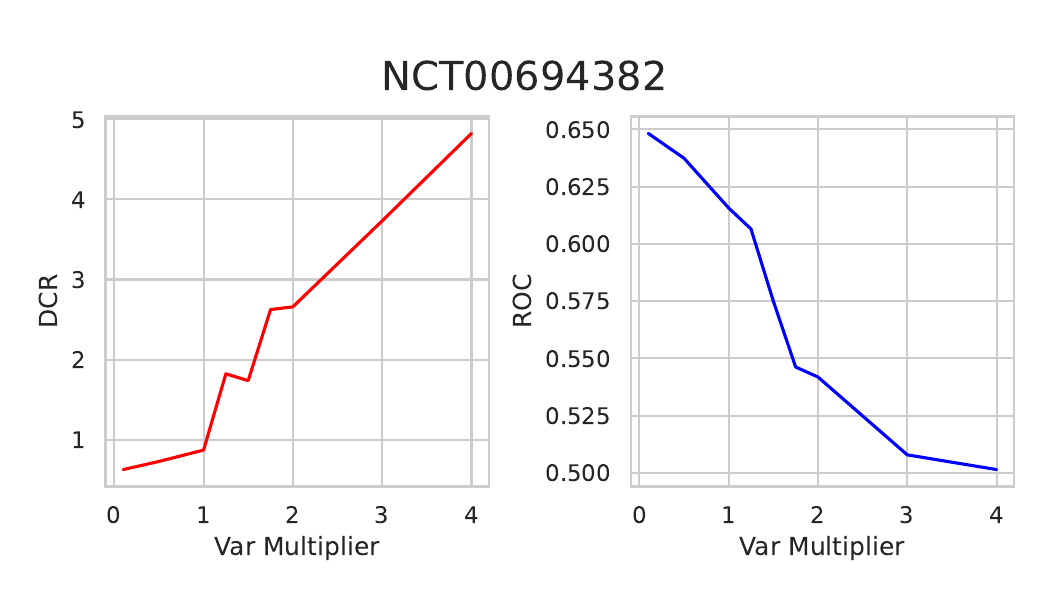}
    \includegraphics[width=0.49\linewidth]{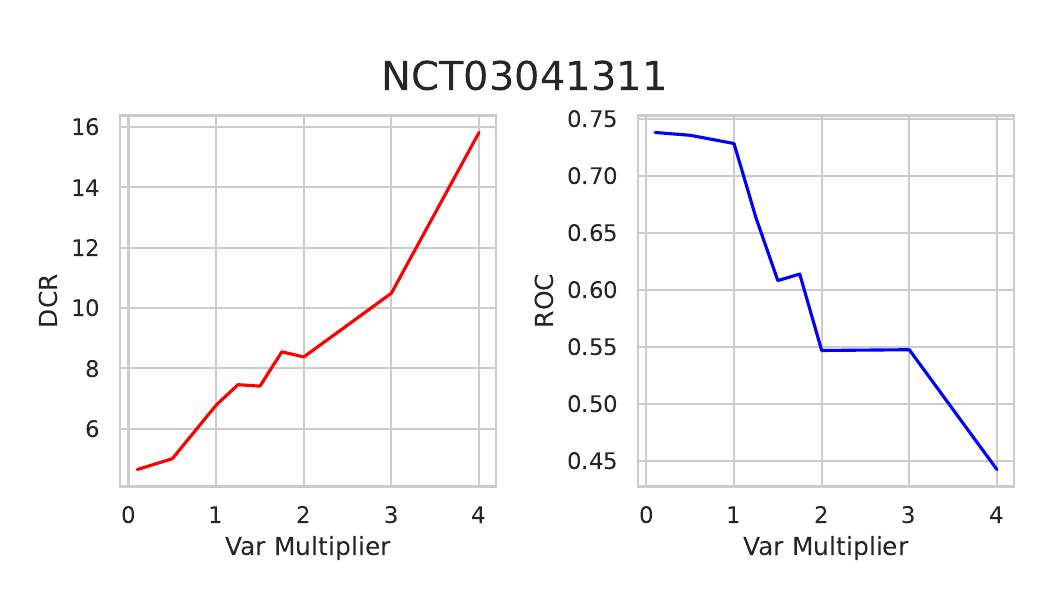}
    \caption{Additional Privacy / Utility tradeoffs examples in \method: Performance of distance to closest record (DCR) (red) and downstream ROC (blue) metrics at varying levels of VAE sampling variance (from 0.1 to 4), represented as the ``Var Multiplier.''}
    \label{fig:add_ablation}
\end{figure}

In this section, we include additional ablations on varying the multiplier on the standard deviation predicted by \method (Events Unknown), shown in Figure~\ref{fig:add_ablation}.

\subsection{Utility / Privacy Spider Plots}
Here, we visualize the utility/privacy trade-off that is inherent to any synthetic data generation task. Each metric is normalized for ease of visualization so that the maximum achieved metric is set as the tip of the triangle by dividing by the max. For ML Inference Privacy (where 0.5 is the ideal value), we first take the absolute value of the difference (i.e. $x = |x-0.5|$), and then divide by the max as before. 

\begin{figure}[ht]
    \centering
    \includegraphics[width=.33\linewidth, trim={0 3cm 0 2cm},clip]{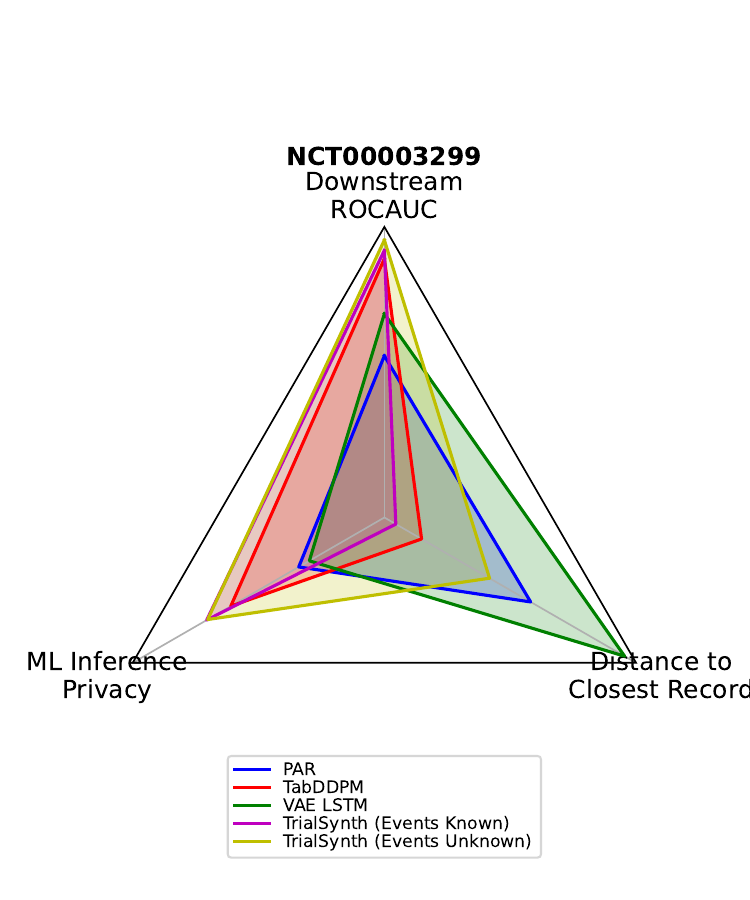}
    \includegraphics[width=.33\linewidth, trim={0 3cm 0 2cm},clip]{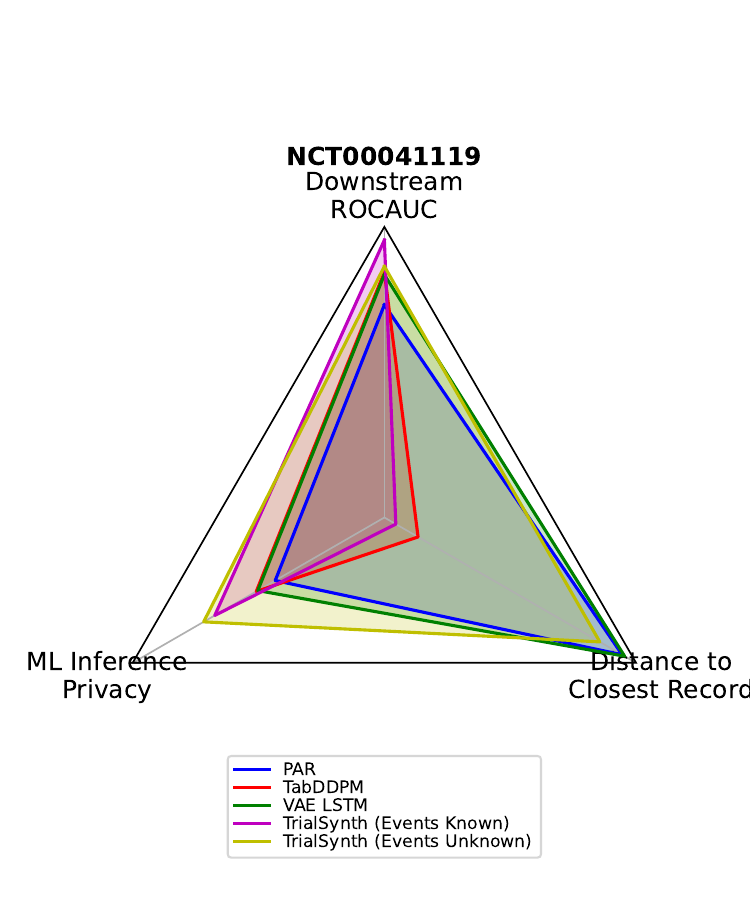}
    \includegraphics[width=.33\linewidth, trim={0 3cm 0 2cm},clip]{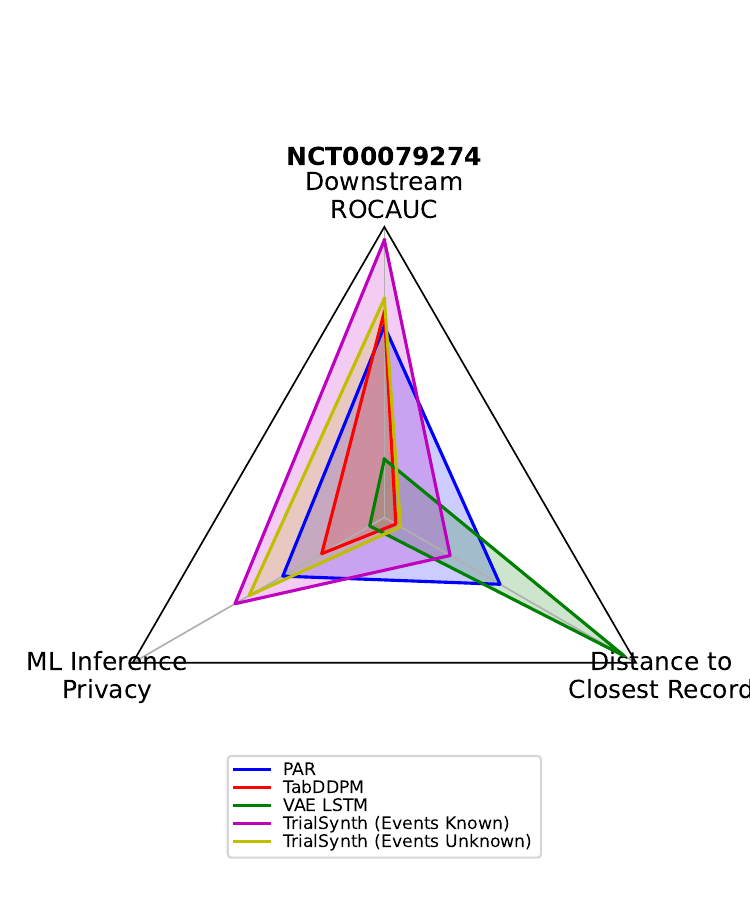}
    \includegraphics[width=.33\linewidth, trim={0 3cm 0 2cm},clip]{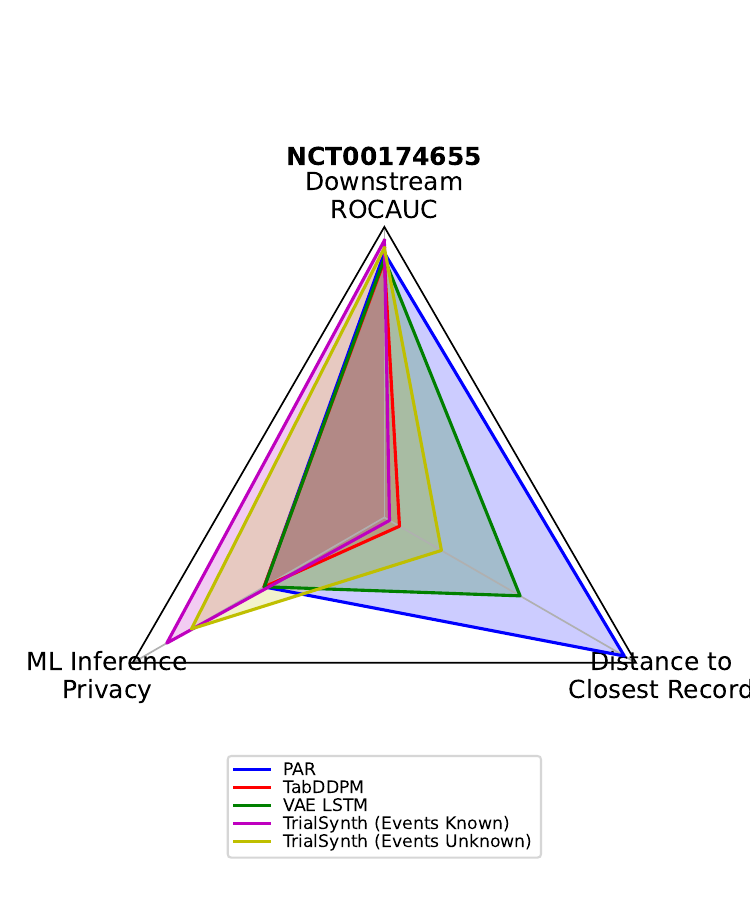}
    \includegraphics[width=.33\linewidth, trim={0 3cm 0 2cm},clip]{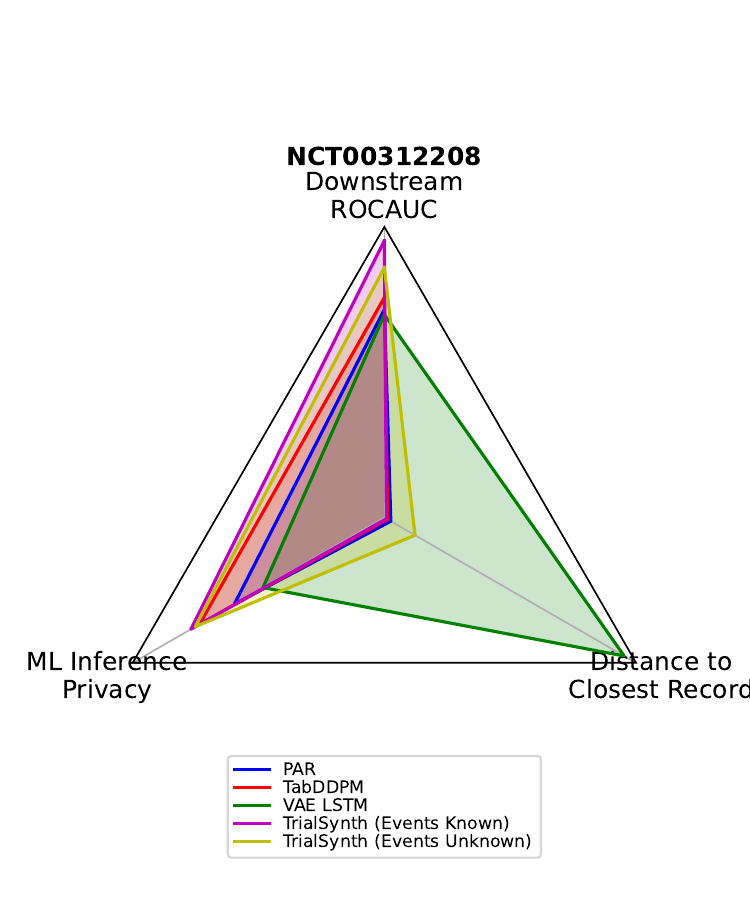}
    \includegraphics[width=.33\linewidth, trim={0 3cm 0 2cm},clip]{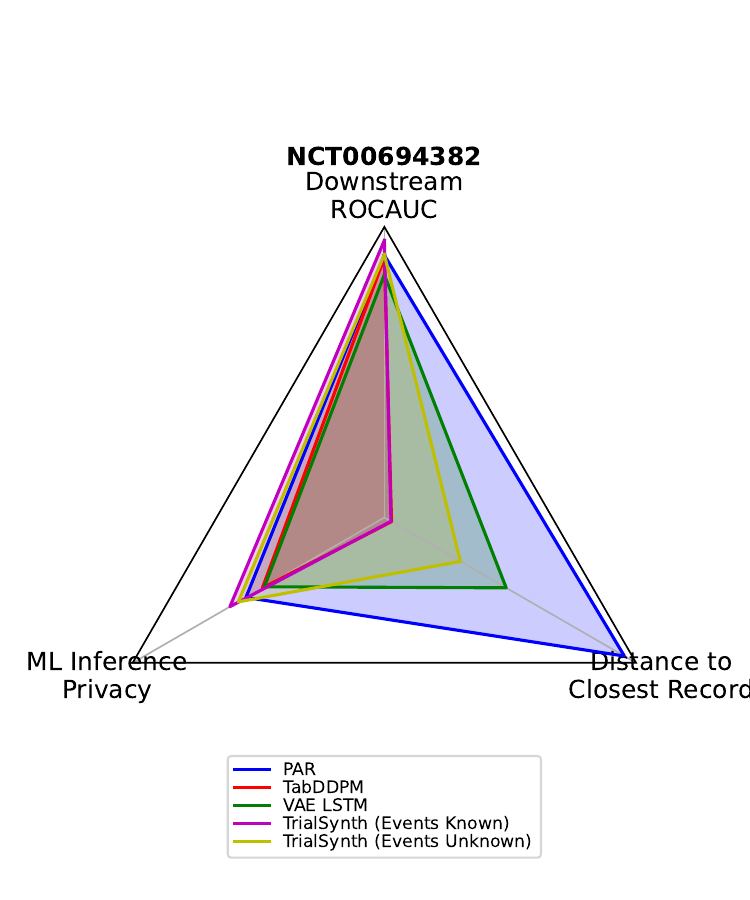}
    \includegraphics[width=.33\linewidth, trim={0 3cm 0 2cm},clip]{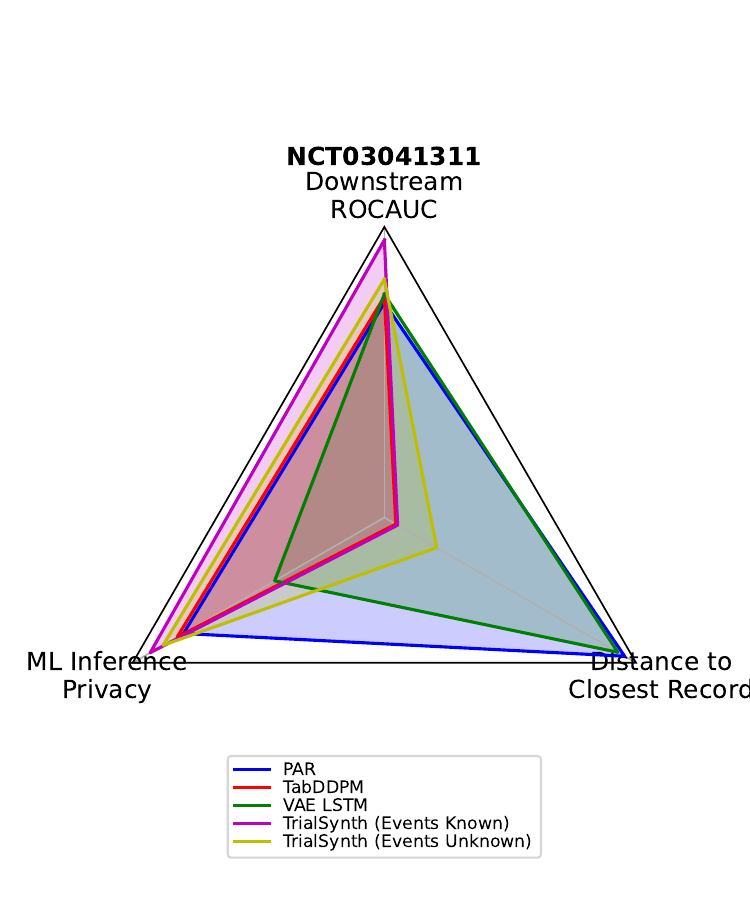}
    \includegraphics[width=.33\linewidth, trim={3cm 0cm 3cm 12cm},clip]{imgs/spider_plot_NCT03041311.pdf}
    \caption{Spider Plots of all Models over all datasets. 
    }
    \label{fig:spiderplots2}
\end{figure}

The results are shown in Figure~\ref{fig:spiderplots2}. We see a clear trade-off, as the best-performing Distance to Closest Record model, usually VAE LSTM or PAR, performs worse on the downstream ROCAC metric. This is because the generated sequences are of poorer quality, being too different from the original. The best-performing Downstream ROCAUC models also generally have good ML Inference Privacy, which is to be expected as those models generate data that is similar to the original, which would allow for (1) better performance on the held-out test set for ROCAUC and (2) being harder to distinguish from original data.

\begin{figure}[ht]
    \centering
    \includegraphics[width=.25\linewidth, trim={.5cm .5cm .5cm 11cm},clip]{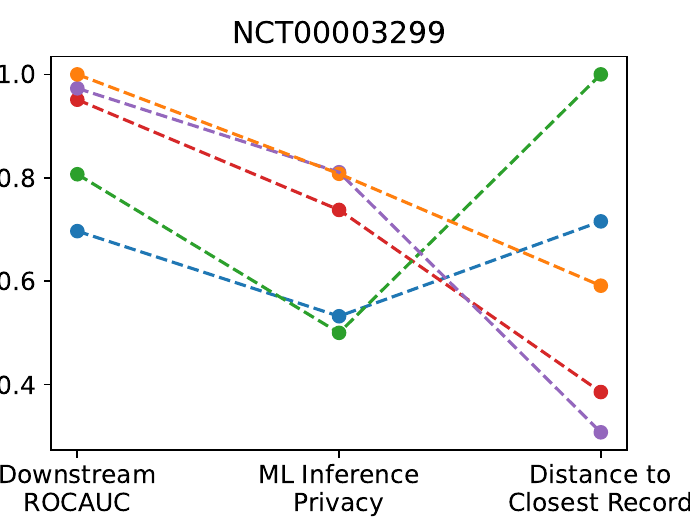}
    \includegraphics[width=.25\linewidth, trim={.5cm .5cm .5cm 11cm},clip]{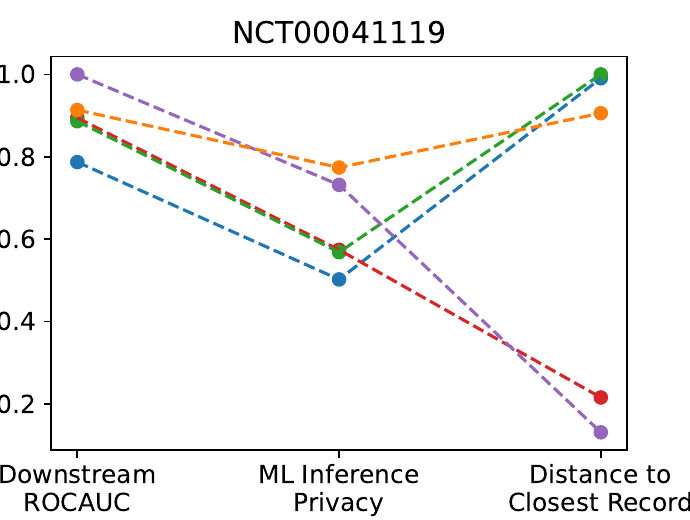}
    \includegraphics[width=.25\linewidth, trim={.5cm .5cm .5cm 11cm},clip]{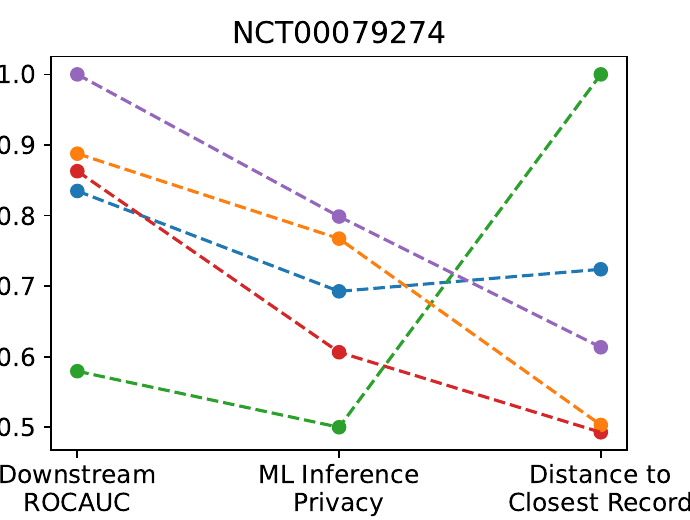}
    \includegraphics[width=.22\linewidth, trim={2cm 10.2cm 2cm 5cm},clip]{imgs/line_plot_NCT00003299.pdf}
    \caption{Line plots of all Models over 3 datasets. All metrics are normalized to scale between 0 and 1. The ROCAUC results are the fidelity results on downstream utility (ML classification of binary patient death/survival). Additional graphs are in the appendix in Figure \ref{fig:spiderplots2}.
}
    \label{fig:spiderplots}
\end{figure}


\newpage
\section*{NeurIPS Paper Checklist}

The checklist is designed to encourage best practices for responsible machine learning research, addressing issues of reproducibility, transparency, research ethics, and societal impact. Do not remove the checklist: {\bf The papers not including the checklist will be desk rejected.} The checklist should follow the references and follow the (optional) supplemental material.  The checklist does NOT count towards the page
limit. 

Please read the checklist guidelines carefully for information on how to answer these questions. For each question in the checklist:
\begin{itemize}
    \item You should answer \answerYes{}, \answerNo{}, or \answerNA{}.
    \item \answerNA{} means either that the question is Not Applicable for that particular paper or the relevant information is Not Available.
    \item Please provide a short (1–2 sentence) justification right after your answer (even for NA). 
\end{itemize}

{\bf The checklist answers are an integral part of your paper submission.} They are visible to the reviewers, area chairs, senior area chairs, and ethics reviewers. You will be asked to also include it (after eventual revisions) with the final version of your paper, and its final version will be published with the paper.

The reviewers of your paper will be asked to use the checklist as one of the factors in their evaluation. While "\answerYes{}" is generally preferable to "\answerNo{}", it is perfectly acceptable to answer "\answerNo{}" provided a proper justification is given (e.g., "error bars are not reported because it would be too computationally expensive" or "we were unable to find the license for the dataset we used"). In general, answering "\answerNo{}" or "\answerNA{}" is not grounds for rejection. While the questions are phrased in a binary way, we acknowledge that the true answer is often more nuanced, so please just use your best judgment and write a justification to elaborate. All supporting evidence can appear either in the main paper or the supplemental material, provided in appendix. If you answer \answerYes{} to a question, in the justification please point to the section(s) where related material for the question can be found.

IMPORTANT, please:
\begin{itemize}
    \item {\bf Delete this instruction block, but keep the section heading ``NeurIPS paper checklist"},
    \item  {\bf Keep the checklist subsection headings, questions/answers and guidelines below.}
    \item {\bf Do not modify the questions and only use the provided macros for your answers}.
\end{itemize}


\begin{enumerate}

\item {\bf Claims}
    \item[] Question: Do the main claims made in the abstract and introduction accurately reflect the paper's contributions and scope?
    \item[] Answer: \answerYes{} 
    \item[] Justification: See conclusion and experiments

\item {\bf Limitations}
    \item[] Question: Does the paper discuss the limitations of the work performed by the authors?
    \item[] Answer: \answerYes{} 
    \item[] Justification: See limitations

\item {\bf Theory Assumptions and Proofs}
    \item[] Question: For each theoretical result, does the paper provide the full set of assumptions and a complete (and correct) proof?
    \item[] Answer: \answerYes{} 
    \item[] Justification: We point to the proof in the original paper and describe it in the appendix

    \item {\bf Experimental Result Reproducibility}
    \item[] Question: Does the paper fully disclose all the information needed to reproduce the main experimental results of the paper to the extent that it affects the main claims and/or conclusions of the paper (regardless of whether the code and data are provided or not)?
    \item[] Answer: \answerNo{} 
    \item[] Justification: Code will be cleaned and anonymised first, before release

\item {\bf Open access to data and code}
    \item[] Question: Does the paper provide open access to the data and code, with sufficient instructions to faithfully reproduce the main experimental results, as described in supplemental material?
    \item[] Answer: \answerYes{} 
    \item[] Justification: Code will be cleaned and anonymised for release

\item {\bf Experimental Setting/Details}
    \item[] Question: Does the paper specify all the training and test details (e.g., data splits, hyperparameters, how they were chosen, type of optimizer, etc.) necessary to understand the results?
    \item[] Answer: \answerYes{} 
    \item[] Justification: Hyperparameters are described, but code will be released soon after.

\item {\bf Experiment Statistical Significance}
    \item[] Question: Does the paper report error bars suitably and correctly defined or other appropriate information about the statistical significance of the experiments?
    \item[] Answer: \answerYes{}{} 
    \item[] Justification: Yes, reported in table captions

\item {\bf Experiments Compute Resources}
    \item[] Question: For each experiment, does the paper provide sufficient information on the computer resources (type of compute workers, memory, time of execution) needed to reproduce the experiments?
    \item[] Answer: \answerYes{} 
    \item[] Justification: See appendix (hyperparameters)
    
\item {\bf Code Of Ethics}
    \item[] Question: Does the research conducted in the paper conform, in every respect, with the NeurIPS Code of Ethics \url{https://neurips.cc/public/EthicsGuidelines}?
    \item[] Answer: \answerYes{} 
    \item[] Justification: Reviewed

\item {\bf Broader Impacts}
    \item[] Question: Does the paper discuss both potential positive societal impacts and negative societal impacts of the work performed?
    \item[] Answer: \answerYes{} 
    \item[] Justification: See limitations
    
\item {\bf Safeguards}
    \item[] Question: Does the paper describe safeguards that have been put in place for the responsible release of data or models that have a high risk for misuse (e.g., pretrained language models, image generators, or scraped datasets)?
    \item[] Answer: \answerYes{} 
    \item[] Justification: Datasets are described, and code will be released
    \item[] Guidelines:
    \begin{itemize}
        \item The answer NA means that the paper poses no such risks.
        \item Released models that have a high risk for misuse or dual-use should be released with necessary safeguards to allow for controlled use of the model, for example by requiring that users adhere to usage guidelines or restrictions to access the model or implementing safety filters. 
        \item Datasets that have been scraped from the Internet could pose safety risks. The authors should describe how they avoided releasing unsafe images.
        \item We recognize that providing effective safeguards is challenging, and many papers do not require this, but we encourage authors to take this into account and make the best faith effort.
    \end{itemize}

\item {\bf Licenses for existing assets}
    \item[] Question: Are the creators or original owners of assets (e.g., code, data, models), used in the paper, properly credited and are the license and terms of use explicitly mentioned and properly respected?
    \item[] Answer: \answerYes{} 
    \item[] Justification: Everything should be correctly cited

\item {\bf New Assets}
    \item[] Question: Are new assets introduced in the paper well documented and is the documentation provided alongside the assets?
    \item[] Answer: \answerNA{} 
    \item[] Justification: Code will be released soon, datasets are described in the appendix

\item {\bf Crowdsourcing and Research with Human Subjects}
    \item[] Question: For crowdsourcing experiments and research with human subjects, does the paper include the full text of instructions given to participants and screenshots, if applicable, as well as details about compensation (if any)? 
    \item[] Answer: \answerNA{} 
    \item[] Justification: We did not collect human subjects data

\item {\bf Institutional Review Board (IRB) Approvals or Equivalent for Research with Human Subjects}
    \item[] Question: Does the paper describe potential risks incurred by study participants, whether such risks were disclosed to the subjects, and whether Institutional Review Board (IRB) approvals (or an equivalent approval/review based on the requirements of your country or institution) were obtained?
    \item[] Answer: \answerNA{} 
    \item[] Justification: We did not collect human subjects data

\end{enumerate}

\end{document}